\documentclass[sigconf,nonacm]{acmart}

\AtBeginDocument{%
  }

\usepackage{multirow}
\usepackage{enumitem}
\usepackage{tcolorbox}
\tcbuselibrary{skins,breakable,listings}
\usepackage{pifont}
\usepackage{placeins}

\newcommand{\cmark}{\ding{51}}
\newcommand{\xmark}{\ding{55}}

\newtcblisting{promptbox}{
  colback=gray!5!white,
  colframe=gray!50!black,
  listing only,
  listing options={
    basicstyle=\ttfamily\scriptsize,
    breaklines=true,
    columns=fullflexible,
    breakindent=0pt,
    breakautoindent=false
  },
  arc=1mm,
  boxrule=0.5pt,
  left=4pt,right=4pt,top=4pt,bottom=4pt,
  breakable
}

\begin{document}

\title[Defake-o3]{Defake-o3: From Speculative Rationales to Verifiable Evidence for Explainable AIGI Detection}

\author{Bowen Deng}
\authornote{Also with Ant Group.}
\orcid{0009-0001-4995-7888}
\email{des\_wv@sjtu.edu.cn}
\affiliation{%
  \institution{Shanghai Jiao Tong University}
  \city{Shanghai}
  \country{China}
}

\author{Jiahui Zhan}
\orcid{0009-0005-6721-6740}
\email{jiahuizhan@sjtu.edu.cn}
\affiliation{%
  \institution{Shanghai Jiao Tong University}
  \city{Shanghai}
  \country{China}
}

\author{Yikun Ji}
\orcid{0000-0001-7788-2361}
\email{da-kun@sjtu.edu.cn}
\affiliation{%
  \institution{Shanghai Jiao Tong University}
  \city{Shanghai}
  \country{China}
}

\author{Haozhen Yan}
\orcid{0009-0001-2223-8402}
\email{orion810@sjtu.edu.cn}
\affiliation{%
  \institution{Shanghai Jiao Tong University}
  \city{Shanghai}
  \country{China}
}

\author{Jianfu Zhang}
\authornotemark[1]
\authornote{Corresponding author.}
\orcid{0000-0002-2673-5860}
\email{c.sis@sjtu.edu.cn}
\affiliation{%
  \institution{Shanghai Jiao Tong University}
  \city{Shanghai}
  \country{China}
}

\settopmatter{authorsperrow=5}
\renewcommand{\shortauthors}{Bowen Deng, Jiahui Zhan, Yikun Ji, Haozhen Yan, and Jianfu Zhang}

\begin{abstract}
The rapid progress of image generation models calls for AI-generated image (AIGI) detectors that are not only accurate but also explainable and reliable. While MLLM-based detectors can provide natural language explanations, existing methods often generate speculative rationales: they rely on vague or hallucinated artifacts, miss subtle localized flaws from the latest generators, and fail to provide evidence that can be visually verified. We present Defake-o3, an explainable AIGI detector that moves from speculative rationales to verifiable evidence. It combines interactive visual search with verifier-guided evidence alignment: the model iteratively zooms into suspicious regions to inspect fine-grained details, while an Evidence Verifier, trained from human verification annotations, provides reinforcement learning rewards that favor grounded evidence and penalize baseless claims. To support this objective, we construct GroundFake, a dataset designed for grounded explainable detection, with localized bounding-box evidence, human verification based on visual grounding and artifact specificity, corrected reasoning trajectories, and valid/invalid evidence supervision. We further introduce FakeFrontier, an out-of-distribution benchmark built from real images and outputs of 10 recent generators, together with an MLLM-based protocol for evaluating evidence quality and persuasiveness. Experiments on GroundFake, FakeFrontier, and additional out-of-distribution benchmarks show that Defake-o3 improves both detection accuracy and explanation quality, producing more localized, verifiable, and persuasive evidence.
\end{abstract}

\maketitle

\begin{figure}[t!]
  \centering
  \includegraphics[width=0.95\linewidth]{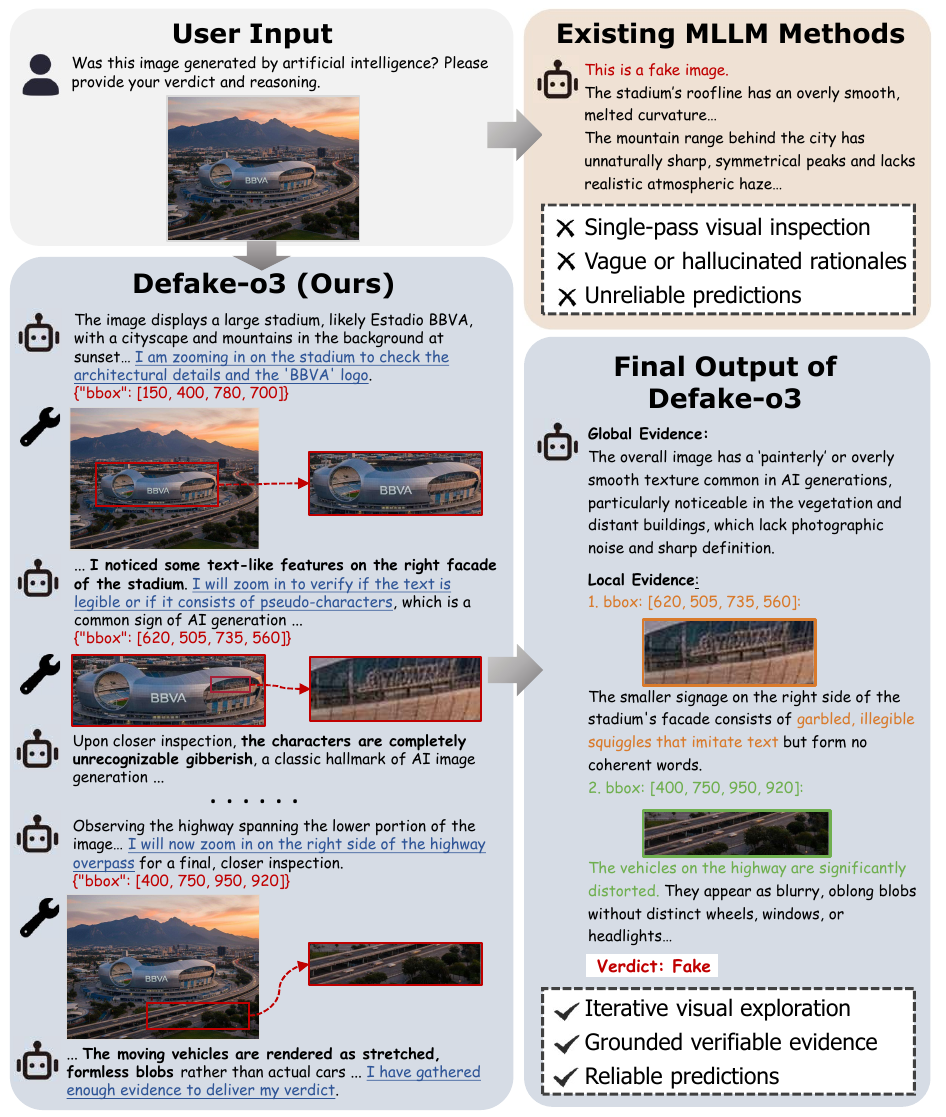}
  \caption{Overview of Defake-o3. Compared with existing single-pass MLLM detectors, Defake-o3 iteratively zooms into suspicious regions to verify candidate artifacts and produces a structured verdict supported by global evidence and localized key evidence.}
  \Description{A comparison of existing single-pass MLLM detectors and Defake-o3. Existing methods inspect the full image once and may produce speculative rationales. Defake-o3 repeatedly selects and magnifies suspicious regions, verifies localized artifacts, and returns an authenticity verdict supported by global evidence and bounding-box-level local evidence.}
  \label{fig:defake_o3_overview}
  \vspace{-12pt}
\end{figure}

\section{Introduction}
\label{sec:intro}
The rapid progress of image generation models, including autoregressive text-to-image systems~\cite{vqvae,imagegpt,delle,parti} and diffusion-based approaches~\cite{ddpm,ddim,ldm,rectifiedflow,imagen,dalle3,stablediffusion3,qwenimage,zimage}, has made synthetic images increasingly photorealistic and harder to distinguish from authentic photographs. Although these advances benefit creative applications, they also create serious risks for public trust by enabling the spread of visually convincing misinformation at scale. As a result, robust and trustworthy AI-generated image (AIGI) detection has become increasingly important. Traditional AIGI detection methods typically cast the task as binary classification. Built on Convolutional Neural Networks (CNNs)~\cite{cnn} and Vision Transformers~\cite{vit}, these data-driven detectors have achieved strong detection performance~\cite{cnnspot,wang2023dire,npr,drct,aide,dda}, yet they remain fundamentally black-box models. In most cases, they output only an authenticity score without revealing the visual evidence behind the decision. As a result, their predictions are difficult to verify, which limits their trustworthiness and practical deployment in high-stakes settings.
Recent studies have explored Multimodal Large Language Models (MLLMs)~\cite{gpt3,flamingo,blip2,instructblip,llava,qwen3vl} for explainable AIGI detection~\cite{fakebench,loki,fakeclue,legion,fakeshield}, producing natural-language rationales alongside authenticity predictions. However, these rationales are often \textbf{speculative} rather than supported by \textbf{verifiable evidence}. First, they are frequently \textit{ungrounded}, relying on vague, generic, or hallucinated cues instead of precise visual artifacts, with no explicit criterion for valid evidence. Second, existing methods generally perform \textit{single-pass, coarse-grained} inspection. As modern generators leave increasingly subtle and localized flaws, standard open-source MLLMs, constrained by input resolution and single-pass visual encoding, often fail to examine suspicious regions in sufficient detail. Consequently, their explanations may appear plausible while remaining weakly grounded, poorly localized, and difficult to verify.

To address these limitations, we propose \textbf{Defake-o3}, an explainable AIGI detector that moves from speculative rationales to verifiable evidence through two complementary components: \textbf{interactive visual search}, which determines where and at what granularity to inspect, and \textbf{verifier-guided evidence alignment}, which defines what constitutes valid evidence. Defake-o3 adopts an iterative ``thinking-with-images'' mechanism~\cite{vstar,zheng2025deepeyes,pixelreasoner}, using a \textbf{zoom-in} tool to crop and magnify suspicious regions and uncover subtle artifacts missed in the full image (Fig.~\ref{fig:defake_o3_overview}). To support this objective, we construct \textbf{GroundFake}, a debiased training dataset that controls image format, aspect ratio, semantic category, and aesthetic quality to reduce shortcut bias. It provides localized bounding-box evidence under strict human verification based on \textit{visual grounding} and \textit{artifact specificity}, together with corrected reasoning trajectories and valid/invalid evidence labels. These annotations support both supervised fine-tuning and the training of an \textbf{Evidence Verifier}, which serves as the reward model~\cite{rlhf} during GRPO-based reinforcement learning~\cite{ppo,grpo,dapo} to reward grounded evidence and penalize baseless claims. For evaluation on recent generators, we introduce \textbf{FakeFrontier}, an out-of-distribution benchmark containing 2,000 real images and 2,000 synthetic images generated by 10 recent models, including Nano~Banana~\cite{nanobanana} and Seedream~4.5~\cite{seedream45}. We also develop an MLLM-based protocol to evaluate both classification performance and the quality and persuasiveness of the produced evidence.

In summary, our contributions are threefold:
\begin{itemize}[leftmargin=*]
\item We propose \textbf{Defake-o3}, an explainable AIGI detector that combines interactive visual search with verifier-guided evidence alignment.
Iterative zoom-in inspection reveals subtle artifacts, while the Evidence Verifier guides reinforcement learning toward grounded evidence and away from baseless claims.
\item We introduce \textbf{GroundFake}, a debiased training dataset for grounded explainable AIGI detection. It provides localized evidence verified according to explicit visual-grounding and artifact-specificity criteria, together with corrected reasoning trajectories and valid/invalid evidence labels for supervised fine-tuning and verifier training.
\item We present \textbf{FakeFrontier}, an out-of-distribution benchmark containing real images and outputs from 10 recent generators, along with an MLLM-based protocol for evaluating both detection performance and evidence quality.
\end{itemize}


\section{Related Work}
\noindent\textbf{Conventional Black-Box AIGI Detection:}
A dominant line of AIGI detection formulates the task as binary classification. CNNSpot~\cite{cnnspot} uses data augmentation to improve cross-model generalization of a standard ResNet detector. DIRE~\cite{wang2023dire} and DRCT~\cite{drct} exploit reconstruction discrepancies revealed by pre-trained diffusion models. NPR~\cite{npr} identifies synthetic images through local pixel statistics introduced by upsampling, while AIDE~\cite{aide} combines frequency and semantic cues in a dual-stream architecture. Recent studies~\cite{dda} further suggest that detector generalization is also sensitive to distributional bias between real and synthetic data.
These methods have established strong classification baselines for AIGI detection, but they remain fundamentally black-box detectors. In practice, they typically output only an authenticity score, without revealing which visual cues support the prediction or providing localized, visually verifiable evidence that humans can inspect. Consequently, their decisions are difficult to interpret and verify, which motivates recent efforts toward explainable AIGI detection.

\noindent\textbf{Explainable Detection via MLLMs:}
To move beyond score-only black-box detectors, recent studies have extended Multimodal Large Language Models (MLLMs) to explainable AIGI detection, while related works also consider broader image or video forgery settings. Instruction-tuning-based methods such as FakeVLM~\cite{fakeclue} and FakeScope~\cite{fakescope} construct large-scale visual instruction data so that MLLMs can jointly predict authenticity and generate textual explanations. IVY-FAKE~\cite{ivyfake} further unifies explainable detection across images and videos. To improve spatial grounding, methods such as LEGION~\cite{legion} and FakeShield~\cite{fakeshield} augment MLLMs with artifact localization, often by leveraging external models such as SAM~\cite{sam} to predict masks over suspicious artifact regions. More recently, alignment strategies based on preference optimization or reinforcement learning have been explored to improve explanation quality. AIGI-Holmes~\cite{aigiholmes} adopts DPO~\cite{dpo} to suppress low-quality explanations, but its outputs remain coarse-grained without explicit localization. So-Fake~\cite{sofake} and FakeXplain~\cite{fakexplain} further combine explanation generation with bounding-box grounding under RL-style training.
However, current methods still mainly optimize the similarity of the explanation with reference annotations, rather than explicitly learning whether a claimed piece of evidence is visually grounded and artifact-specific enough to support an AIGI verdict. As a result, even localized outputs can remain vague, spurious, or hallucinated, and existing resources still provide limited evidence-level supervision for training and evaluating truly verifiable explanations.

\begin{figure*}[ht]
  \centering
  \includegraphics[width=0.95\textwidth]{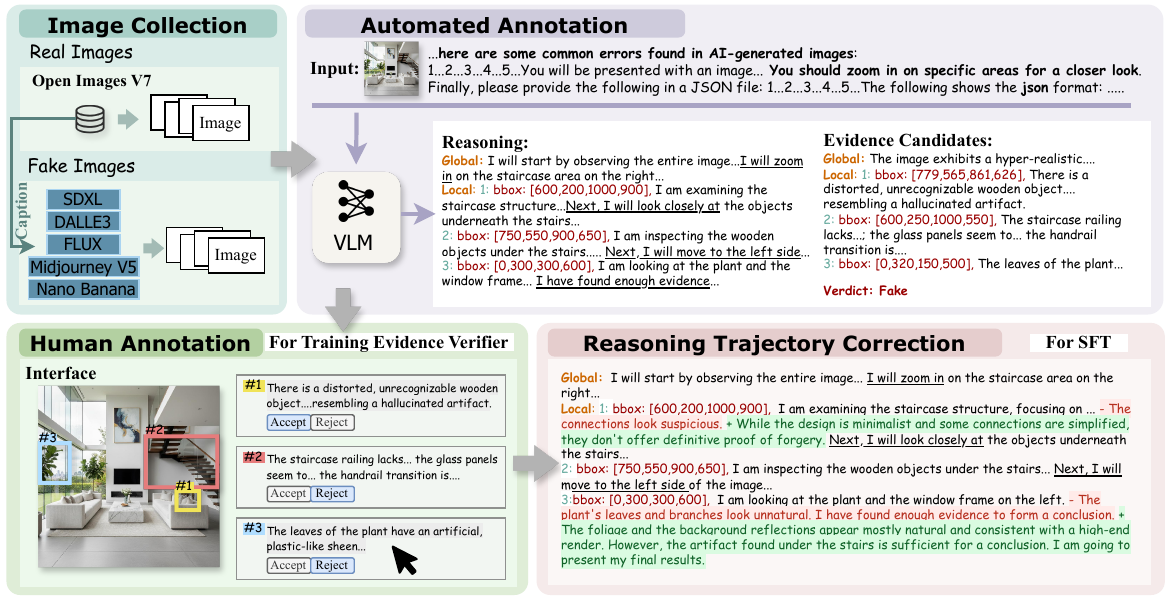}
  \caption{GroundFake construction pipeline. Starting from a balanced, bias-controlled image pool, we generate label-conditioned candidate reasoning trajectories and localized evidence, manually verify localized key evidence for fake images, and rewrite the trajectories into annotation-consistent supervision transcripts. The resulting dataset supports both SFT and Evidence Verifier training.}
  \Description{A four-stage pipeline for constructing GroundFake: collecting a balanced and bias-controlled image pool, generating candidate reasoning and localized evidence, manually verifying the localized evidence, and correcting reasoning trajectories for SFT and Evidence Verifier training.}
  \label{fig:data_pipeline}
\end{figure*}

\noindent\textbf{Dynamic Visual Reasoning and Exploration:}
Recent advances in MLLM-based visual reasoning have shifted from fixed, text-dominant chain-of-thought toward interactive visual exploration, where models actively decide where to inspect and at what granularity. Early works such as Visual CoT~\cite{visualcot} and LLaVA-CoT~\cite{llavacot} made reasoning more explicit through region grounding or staged decomposition, but still relied largely on static visual inputs. Later studies treated visual search as a core reasoning primitive: V*~\cite{vstar} introduced LLM-guided search for high-resolution detail grounding, Visual-RFT~\cite{visualrft} showed that verifiable rewards can improve visual reasoning behavior, and Pixel Reasoner~\cite{pixelreasoner}, DeepEyes~\cite{zheng2025deepeyes}, and Mini-o3~\cite{lai2025mini} further enabled models to zoom, inspect, and iteratively interact with images during inference. 
These advances are especially relevant to explainable AIGI detection, where recent generators leave increasingly subtle and localized artifacts that are hard to capture with single-pass visual encoding. However, they mainly improve where to look, rather than what should count as valid evidence for an AIGI verdict. Therefore, interactive visual search alone does not guarantee grounded, artifact-specific, and visually verifiable explanations.

\section{GroundFake Dataset Construction}
To support grounded explainable AIGI detection, we construct\\ \textbf{GroundFake}, a training dataset containing 16,000 images, stepwise reasoning transcripts, and evidence annotations. For fake images, the localized key evidence is manually verified under visual grounding and artifact specificity criteria. We further rewrite the reasoning trajectories using the verified evidence for annotation consistency. GroundFake provides two complementary supervision signals for Defake-o3: rewritten trajectories for supervised fine-tuning and valid/invalid evidence labels for Evidence Verifier training. The overall construction pipeline of GroundFake is illustrated in Fig.~\ref{fig:data_pipeline}.

\subsection{Image Collection and Bias Control}
We construct a balanced and bias-controlled image pool of 16k images (8k real and 8k fake) for GroundFake. Real images are sampled from Open Images V7~\cite{openimagev7}, while fake images are collected from recent image generators and supplemented with additional FLUX.1-dev samples~\cite{flux1dev} to broaden photographic conditions. To reduce dataset-specific shortcuts, we standardize file format and align the aspect-ratio, coarse semantic-category, and aesthetic-quality distributions between the real and fake subsets, encouraging the detector to rely on generation artifacts rather than trivial source cues.

  

\subsection{Label-Conditioned Candidate Annotation}

At this stage, we use Gemini 3 Pro Preview~\cite{gemini3pro} to produce candidate reasoning trajectories and evidence proposals conditioned on the known image label. The ground-truth label is provided only to elicit class-consistent candidate evidence and localization proposals. The prompt lists several common artifact categories in AI-generated images, including incoherent text, deformed faces or hands, distorted object structures, and violations of common sense or physical constraints, as heuristic cues for candidate generation. The model is then instructed to inspect the image region by region, with attention to fine-grained details, and to return a structured output containing a reasoning trajectory, a label-consistent conclusion, and supporting evidence.
Specifically, the generated output contains three components:
\begin{itemize}[leftmargin=*]
    \item \textbf{Reasoning Trajectory:} 
    A sequential list in which each step records a region of interest (specified by a bounding box) together with the corresponding intermediate reasoning text.
    \item \textbf{Verdict:} 
    A label-consistent conclusion in \{``real'', ``fake''\}, included to maintain a complete supervision format.
    \item \textbf{Evidence:} 
    The visual and textual basis associated with the conclusion. For both real and fake images, the model provides one \textit{global evidence item} that summarizes coarse scene-level coherence. For fake images, it further proposes several localized \textbf{key evidence items}, each paired with a bounding box that points to a candidate generation artifact.
\end{itemize}

We treat the Gemini outputs as candidate annotations. Gemini 3 Pro Preview provides useful candidate trajectories and localization proposals, but a non-trivial portion of the localized key evidence remains weakly grounded or hallucinated. We therefore conduct a subsequent human verification step on the localized key evidence for fake images before these annotations are used for trajectory rewriting or Evidence Verifier training.

\subsection{Human Annotation and Verification}\label{sec:human_verification}



We focus human verification on the localized key evidence for fake images, because these artifact claims are the most hallucination-prone and are the annotations later used for verifier supervision. 
The proposed candidate key evidence may still be weakly grounded or invalid. Typical failure cases include: (1) the evidence text is irrelevant to the content inside the bounding box; (2) the text contradicts the actual visual content (e.g., correctly rendered text or normal hands are falsely described as defective); or (3) the claimed cue is not specific enough to AI generation (e.g., ``smooth skin'' which may also appear in heavily post-processed real photographs).
We therefore ask human annotators to judge each localized key evidence item, consisting of the original image, the bounding box, and the associated text, with a binary label: \textit{Valid} or \textit{Invalid}. Each item is independently reviewed by three annotators under two criteria:

\begin{enumerate}[leftmargin=*]
    \item \textbf{Visual Grounding:} 
    The textual description accurately matches the visual content within the bounding box.
    \item \textbf{Artifact Specificity:} 
    The described flaw is more specific to AI generation than a generic attribute that may also appear in real photographs.
\end{enumerate}

We retain the vote ratio across the three annotators as a soft validity label for Evidence Verifier training, and use the retained valid evidence in the subsequent trajectory rewriting step. This yields a higher-quality subset of localized evidence by filtering out unreasonable, weakly grounded, or non-specific indicators of AIGI.

\subsection{Reasoning Trajectory Rewriting for Annotation Consistency}
Simply discarding invalid key evidence would leave the reasoning transcript inconsistent with the retained evidence annotations used for training. We therefore employ Gemini 3 Flash Preview~\cite{geminiflash} to rewrite the reasoning trajectories.
Given the original trajectory and the human-verified valid evidence list, the model rewrites the intermediate reasoning text so that the resulting transcript remains consistent with the verified evidence and no longer depends on discarded hallucinated claims. The rewritten trajectories are then used as annotation-consistent supervision transcripts, providing stable training targets for tool use and final output formatting.

\begin{figure*}[ht]
  \centering
  \includegraphics[width=0.95\linewidth]{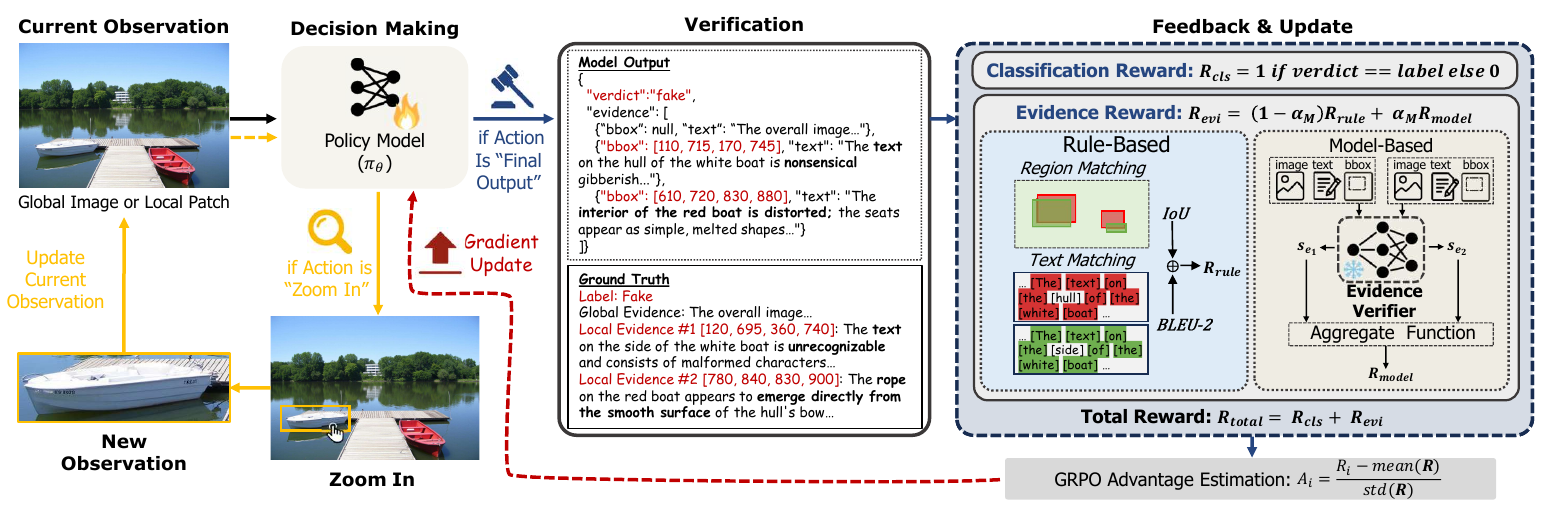}
  \caption{Overview of Defake-o3. The model performs interactive visual search by iteratively zooming into suspicious regions and then outputs a structured verdict with global and localized evidence. For every correctly classified fake image, the total reward combines the classification reward with rule-based evidence matching and model-based scoring from the Evidence Verifier.}
  \Description{The training pipeline of Defake-o3, where the model alternates between viewing the image and selecting zoom-in boxes, then outputs a verdict with localized evidence, and receives GRPO rewards from an evidence verifier.}
  \label{fig:method}
\end{figure*}

\section{Defake-o3 Methodology}
To address the limitations of single-pass and weakly grounded MLLM detectors, we propose \textbf{Defake-o3}. Defake-o3 combines interactive visual search with verifier-guided evidence alignment. At inference time, it iteratively zooms into suspicious regions and then outputs a structured verdict with global evidence and, when available, localized key evidence. Training proceeds in two stages: SFT on human-filtered GroundFake traces to learn tool use and structured evidence generation, followed by GRPO with a hybrid evidence reward that combines rule-based matching and an Evidence Verifier trained from human verification labels. An overview is shown in Fig.~\ref{fig:method}.

\subsection{Task Formulation}\label{sec:formulation}
Given an input image $\mathcal{I}$, the Defake-o3 model $\pi_\theta$ performs a multi-turn interaction with the image and then produces a structured output.

\noindent\textbf{Interactive Visual Search.}
At turn $i$, the model receives an observation $O_i$, which is either the full image or a zoomed-in patch from a previous step, and chooses an action $A_i \in \{\texttt{Zoom In}, \texttt{Final Output}\}$. A \texttt{Zoom In} action predicts a 2D bounding box $b_i$ and returns the corresponding crop as the next observation $O_{i+1}$. A \texttt{Final Output} action terminates the interaction and triggers the final output.

\noindent\textbf{Final Structured Output.}
We denote the final output by $\mathcal{Y} = (\hat{y}, e_g, E)$, where $\hat{y} \in \{\texttt{real}, \texttt{fake}\}$ is the verdict, $e_g = (\emptyset, t_g)$ is a mandatory global evidence item summarizing image-level cues, and $E = \{e_1, \dots, e_K\}$ is a possibly empty set of localized key evidence items. Each $e=(b,t)\in E$ consists of a bounding box $b \neq \emptyset$ and a text description $t$ of a local artifact. For a real verdict, we set $E=\emptyset$. For a fake verdict, the model outputs localized key evidence when clear and visually verifiable local artifacts are found. Otherwise, it may keep $E=\emptyset$ and return only the global evidence to avoid hallucinated localization.
Throughout this paper, we reserve the term verifiable evidence for the localized evidence items in $E$; the global evidence $e_g$ serves only as a contextual image-level assessment.

\subsection{Supervised Fine-Tuning (SFT)}

In the SFT stage, we bootstrap the base MLLM using the human-filtered and trajectory-corrected traces in GroundFake. Each training sample is cast as a multi-turn interaction sequence in which the model learns to invoke the zoom-in tool, condition on the returned patches, and emit the final structured output defined in Sec.~\ref{sec:formulation}. We train the model with standard autoregressive next-token prediction over the full trace. This stage mainly initializes tool use, sequential visual context accumulation, and structured evidence generation, while evidence quality is further aligned in the subsequent RL stage.

\subsection{Evidence Verifier Training}\label{sec:verifier}

For grounded AIGI detection, exact matching to reference annotations, such as spatial overlap with human boxes~\cite{fakexplain,sofake}, is useful but insufficient as a sole reward signal. Reference annotations can be incomplete, and annotation matching alone does not explicitly penalize invalid or fabricated evidence. To provide a denser and more human-aligned signal, we train an \textbf{Evidence Verifier} $V_\phi$ under the human verification protocol in Sec.~\ref{sec:human_verification}. In GroundFake, each localized evidence item is judged by visual grounding and artifact specificity. Using these annotations, the verifier takes the original image $\mathcal{I}$, the cropped local region $\mathcal{I}_b$, the bounding box $b$, and the evidence text $t$ as input, and predicts a validity score $s \in [0,1]$. We instantiate $V_\phi$ by appending a linear classification head to an MLLM and optimize it as a binary classifier with soft labels derived from annotator agreement. Specifically, if $m$ out of $M$ annotators mark an evidence item as valid, we set its target to $y_{\text{soft}} = m/M$ and train the verifier with a soft-label binary cross-entropy loss:
\begin{equation}
s = V_\phi(\mathcal{I}, \mathcal{I}_b, b, t), \quad
\mathcal{L}_{\text{ver}} = - y_{\text{soft}} \log s - (1-y_{\text{soft}})\log(1-s).
\end{equation}
This allows $V_\phi$ to capture the graded certainty of human judgments and provide a learned validity signal for the subsequent RL stage.

\subsection{Reward Function Design}

Our RL objective combines task correctness with localized evidence quality. We penalize unsupported evidence strongly and avoid rewarding the model for proposing many weak evidence items. Accordingly, the reward is defined on the verdict and the localized key evidence set $E$ from Sec.~\ref{sec:formulation}.

\subsubsection{Total Reward ($R_{total}$)}
For a final output $\mathcal{Y}=(\hat{y}, e_g, E)$, where $E=\{e_1,\dots,e_K\}$ denotes the localized key evidence set, we define
\begin{equation}\label{eqn:total}
R_{total}=
\begin{cases}
R_{fail}, & \text{invalid format},\\
R_{cls}+\mathbb{I}(\hat{y}=\texttt{fake})\,R_{evi}, & \text{otherwise},
\end{cases}
\end{equation}
where \emph{invalid format} means that at least one tool call or the final structured output fails to parse, and we set $R_{fail}=-1$. The indicator function $\mathbb{I}(\cdot)$ equals $1$ when its argument is true and $0$ otherwise. 

The classification reward measures only verdict correctness:
\begin{equation}
R_{cls}=\mathbb{I}(\hat{y}=y_{gt}),
\end{equation}
where $y_{gt}$ is the ground-truth label; while the finer-grained shaping of localized evidence quality is handled by $R_{evi}$.

\subsubsection{Evidence Reward ($R_{evi}$)}
We compute the evidence reward on the localized key evidence set $E$. We use a hybrid evidence reward that combines a coarse rule-based matching term with a verifier-based validity term:
\begin{equation}
R_{evi} = (1 - \alpha_M) R_{rule} + \alpha_M R_{model},
\end{equation}
where $R_{rule}$ anchors the policy to the available reference annotations, while $R_{model}$ provides a learned signal for evidence validity under the human verification protocol (see Sec.~\ref{sec:verifier}). This hybrid form avoids relying entirely on either exact annotation matching or the learned verifier alone. Unless otherwise stated, we set $\alpha_M = 0.5$.

\subsubsection{Rule-based Reward ($R_{rule}$)} 
Although exact matching is insufficient as a sole reward signal, it remains a useful coarse anchor when reference localized evidence is available. For the predicted localized evidence set $E$, we form a union mask $M_{pred}$ from all predicted boxes and concatenate all evidence texts into $T_{pred}$. Likewise, for the reference localized evidence, we form $M_{ref}$ and $T_{ref}$. If the model incorrectly classifies a \textit{real} image as \textit{fake}, we set $R_{rule} = 0$. Otherwise, the rule-based reward is defined as:
\begin{equation}
R_{rule} = \lambda_{iou} \text{IoU}(M_{pred}, M_{ref}) + \lambda_{bleu} \text{BLEU-2}(T_{pred}, T_{ref})
\end{equation}
where $\lambda_{iou}=\lambda_{bleu}=1$ by default. 
We use set-level matching, since reference annotations may be incomplete and multiple valid decompositions of evidence can exist for the same image. The BLEU-2 term serves as a lightweight lexical anchor. 

\subsubsection{Model-based Reward ($R_{model}$)} 
For each localized evidence item $e=(b,t)\in E$, we extract the corresponding crop $\mathcal{I}_b$ and query the Evidence Verifier $V_\phi$, obtaining a score $s_e = V_\phi(\mathcal{I}, \mathcal{I}_b, b, t)$.
We then convert the verifier score into an item-level reward:
\begin{equation}
r_{model,e} =
\begin{cases}
-C_{\text{invalid}}, & s_e < \tau,\\
C_{\text{valid}} \cdot \dfrac{s_e - \tau}{1 - \tau}, & s_e \ge \tau,
\end{cases}
\end{equation}
where $\tau$ is the acceptance threshold under the verification protocol in Sec.~\ref{sec:verifier}. This mapping assigns a fixed penalty to low-scoring evidence and only bounded positive reward to evidence whose verifier score exceeds $\tau$. When a real image is incorrectly predicted as fake, the same mapping naturally penalizes the proposed localized evidence, since such evidence is unlikely to pass the verifier threshold. Unless otherwise stated, we use $\tau=0.5$ and $C_{\text{valid}} = C_{\text{invalid}} = 1$.

To avoid rewarding the model for proposing many borderline evidence items, we aggregate item-level rewards with cumulative penalties and capped positive rewards. Let
\begin{equation}
\mathcal{E}^- = \{e \mid r_{model,e} < 0\}, \qquad
\mathcal{E}^+ = \{e \mid r_{model,e} \ge 0\},
\end{equation}
and let $r_{(1)} \ge r_{(2)}$ denote the largest and second-largest reward of the evidence in $\mathcal{E}^+$, respectively. We define:
\begin{equation}
R_{model} = \underbrace{\sum_{e \in \mathcal{E}^-} r_{model,e}}_{\text{Penalty}} + \underbrace{\mathcal{A}(\mathcal{E}^+)}_{\text{Reward}}, \quad
\mathcal{A}(\mathcal{E}^+) =
\begin{cases}
0, & |\mathcal{E}^+| = 0,\\
r_{(1)}, & |\mathcal{E}^+| = 1,\\
r_{(1)} + \beta \, r_{(2)}, & |\mathcal{E}^+| \ge 2.
\end{cases}
\end{equation}
This aggregation reflects our preference for a small number of strong, verifiable artifacts rather than many weak ones. In our implementation, we set $\beta=0.5$.

\subsection{Reinforcement Learning}

Starting from the SFT-trained weights, we optimize Defake-o3 using GRPO~\cite{grpo,dapo}. Each sampled rollout contains multiple rounds of \texttt{Zoom In} actions followed by a final structured output. Through trajectory-level optimization, the model learns to use the zoom-in tool judiciously and to favor a small number of high-quality localized evidence items over many weak or vague ones.

\section{Experiments}
\subsection{Experimental Setup}

\begin{table}[t]
  \caption{Results on the GroundFake test set.}
  \label{tab:groundfake_results}
  \centering
  \scriptsize
  \setlength{\tabcolsep}{3pt}
  \begin{tabular}{@{}lcccccc@{}}
    \toprule
    Method & Acc & F1 & BLEU-1 & BLEU-2 & ROUGE-L & IoU \\
    \midrule
    CNNSpot & 0.976 & 0.976 & -- & -- & -- & -- \\
    NPR & 0.838 & 0.830 & -- & -- & -- & -- \\
    AIDE & 0.924 & 0.922 & -- & -- & -- & -- \\
    \midrule
    FakeVLM & 0.857 & 0.854 & 0.105 & 0.045 & 0.110 & -- \\
    LEGION & 0.502 & 0.024 & 0.003 & 0.001 & 0.002 & 0.000 \\
    FakeShield & 0.494 & 0.275 & 0.029 & 0.013 & 0.026 & 0.017 \\
    \midrule
    Defake-Direct (w/o RL) & 0.946 & 0.943 & 0.282 & 0.154 & 0.234 & 0.165 \\
    Defake-Direct & 0.958 & 0.957 & 0.333 & 0.195 & 0.252 & 0.262 \\
    Defake-CoT (w/o RL) & 0.988 & 0.988 & 0.330 & 0.184 & 0.267 & 0.222 \\
    Defake-CoT & \textbf{0.992} & \textbf{0.992} & 0.335 & 0.184 & 0.257 & 0.280 \\
    Defake-o3 (w/o RL) & 0.972 & 0.971 & 0.319 & 0.179 & 0.250 & 0.241 \\
    \midrule
    \textbf{Defake-o3} & \textbf{0.992} & \textbf{0.992} & \textbf{0.364} & \textbf{0.215} & \textbf{0.281} & \textbf{0.311} \\
    \bottomrule
  \end{tabular}
  
\vspace{1ex}
\raggedright
\footnotesize \centering
Acc/F1: image-level classification performance. BLEU-1/2 and ROUGE-L: evidence-level lexical overlap with the reference annotations. IoU: evidence-level spatial overlap with the reference regions.
\end{table}

\begin{figure*}[t]
  \centering
  \includegraphics[width=0.95\linewidth]{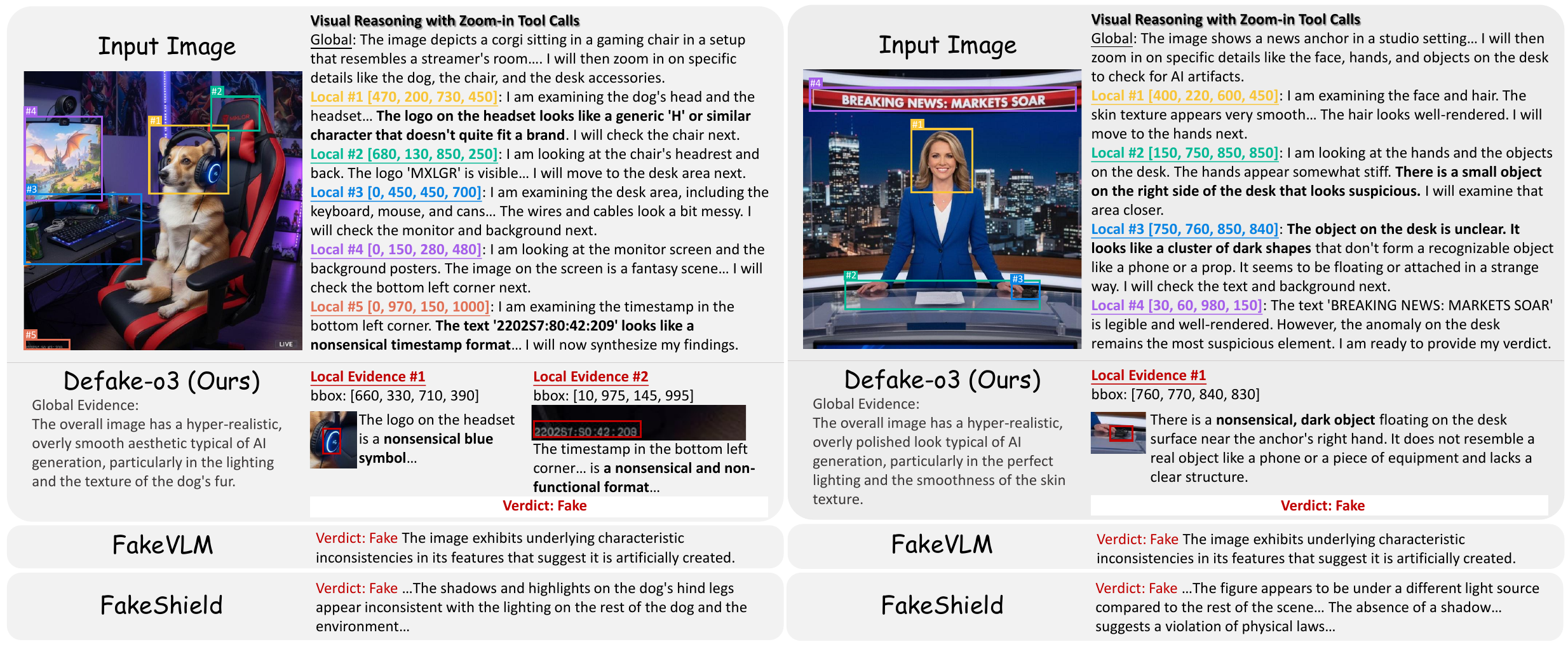}
  \caption{Qualitative examples from the GroundFake test set. Defake-o3 first performs coarse scene-level inspection and then uses targeted zoom-in steps to verify candidate artifacts, yielding a small set of localized, visually verifiable evidence items. Compared with FakeVLM~\cite{fakeclue} and FakeShield~\cite{fakeshield}, its outputs are more directly grounded in explicit, image-specific artifacts. FakeShield masks are omitted for brevity.}
  \Description{Qualitative comparisons on several GroundFake images. For each image, the figure contrasts evidence produced by FakeVLM, FakeShield, and Defake-o3, and shows how Defake-o3 zooms into suspicious regions and identifies localized visual artifacts.}
  \label{fig:showcases}
\end{figure*}

\noindent\textbf{Implementation Details.} 
Unless otherwise stated, all experiments are conducted on 8 A100 GPUs. 
We employ Qwen3-VL-8B-Instruct~\cite{qwen3vl} as our base model and fine-tune it using LoRA~\cite{lora}, which is applied to all linear layers within the text backbone, with a rank of $r=16$ and an alpha of $\alpha=32$. During the SFT stage, we train the model using a global batch size of $8$ and a learning rate of $1 \times 10^{-4}$. In the RL stage, the group size is set to $16$, and the learning rate is $1 \times 10^{-5}$. Following DAPO~\cite{dapo}, we implement several improvements to the standard GRPO algorithm: (1) removing the KL divergence constraint; (2) adopting token-level loss instead of sequence-level loss; (3) applying a clip higher strategy with $\epsilon_{\text{low}}=0.2$ and $\epsilon_{\text{high}}=0.28$; and (4) utilizing overlong reward masking, a mechanism that limits sampling length without suppressing the generation of longer reasoning paths. The Evidence Verifier shares the same base model, LoRA configuration, and SFT training settings, with the exception that its linear classification head is fully trainable.

\noindent\textbf{Baselines and Ablations.} 
We compare Defake-o3 with traditional binary classification approaches including CNNSpot~\cite{cnnspot}, NPR~\cite{npr}, and AIDE~\cite{aide}. For a fair comparison, these models are re-trained on our GroundFake training set. For MLLM-based explainable methods, we employ and test FakeVLM~\cite{fakeclue}, LEGION~\cite{legion}, and FakeShield~\cite{fakeshield}. We evaluate FakeVLM and FakeShield using their officially released weights, while LEGION is reproduced utilizing its official training code and data. 
To validate the effectiveness of our proposed method, we design two variants of our model for ablation studies: 
(1) \textit{Defake-CoT}: relies solely on pure textual chain-of-thought reasoning without zoom-in tool calls; and 
(2) \textit{Defake-Direct}: directly outputs the final verdict and evidence without any preliminary reasoning steps.

\begin{table*}[t]
\centering
\caption{Evaluation with the MLLM-based protocol on the fake-image subset of FakeFrontier.}
\label{tab:fakefrontier_mllm}
\small
\setlength{\tabcolsep}{4pt}
\begin{tabular}{l c ccc ccc ccc}
\toprule
\multirow{2}{*}{\textbf{Method}}
& \multirow{2}{*}{\textbf{Avg \#Evi}}
& \multicolumn{3}{c}{\textbf{Qwen3-VL-235B (Prior-Acc = 0.2421)}}
& \multicolumn{3}{c}{\textbf{Kimi K2.5 (Prior-Acc = 0.3526)}}
& \multicolumn{3}{c}{\textbf{GLM-4.6V (Prior-Acc = 0.0737)}} \\
\cmidrule(lr){3-5} \cmidrule(lr){6-8} \cmidrule(lr){9-11}
& & \makebox[1.2cm][c]{QS} & \makebox[1.2cm][c]{Hit@Img} & \makebox[1.2cm][c]{Hit@Evi} & \makebox[1.2cm][c]{QS} & \makebox[1.2cm][c]{Hit@Img} & \makebox[1.2cm][c]{Hit@Evi} & \makebox[1.2cm][c]{QS} & \makebox[1.2cm][c]{Hit@Img} & \makebox[1.2cm][c]{Hit@Evi} \\
\midrule
\textbf{Defake-o3}             & 3.0804 & \textbf{0.5378} & \textbf{0.7839} & \textbf{0.6362} & \textbf{0.4334} & \textbf{0.7471} & \textbf{0.5895} & \textbf{0.6812} & \textbf{0.9045} & \textbf{0.8249} \\
Defake-CoT                   & 2.9397 & 0.4841 & 0.7755 & 0.6239 & 0.3811 & 0.7119 & 0.5704 & 0.5980 & 0.8961 & 0.7584 \\
Defake-Direct                & 3.1910 & 0.3721 & 0.6047 & 0.4861 & 0.2877 & 0.5645 & 0.4504 & 0.4431 & 0.7085 & 0.5554 \\
\midrule
FakeVLM               & 3.6421 & 0.1268 & 0.3702 & 0.2558 & 0.1204 & 0.4526 & 0.3531 & 0.2438 & 0.5456 & 0.5092 \\
FakeShield            & 2.2161 & 0.0875 & 0.2312 & 0.3296 & 0.0763 & 0.2747 & 0.4218 & 0.1393 & 0.2747 & 0.4255 \\
\bottomrule
\end{tabular}

\vspace{1ex}
\raggedright
\footnotesize
Avg \#Evi: average number of evidence items per image. Prior-Acc: probability that the MLLM evaluator classifies an image as fake without additional information. QS: Quality Score. Hit@Img / Hit@Evi: Image Hit Rate / Evidence Hit Rate. LEGION is excluded from the explainability evaluation, as it classifies nearly all fake images as real.
\end{table*}

\begin{table}[t]
\centering
\caption{Classification results on FakeFrontier benchmark.}
\label{tab:fakefrontier_cls}
\small
\setlength{\tabcolsep}{4pt}
\begin{tabular}{lcccc}
\toprule
\textbf{Method} & \textbf{Real Acc} & \textbf{Fake Acc} & \textbf{Acc} & \textbf{F1} \\
\midrule
\textbf{Defake-o3}             & 0.9020 & 0.9446 & \textbf{0.9232} & \textbf{0.9246} \\
Defake-CoT                   & 0.8675 & 0.9532 & 0.9102 & 0.9137 \\
Defake-Direct                & 0.8485 & 0.7826 & 0.8157 & 0.8088 \\
\midrule
FakeVLM               & 0.7127 & 0.7102 & 0.7114 & 0.7108 \\
FakeShield            & 0.8135 & 0.4107 & 0.6127 & 0.5139 \\
LEGION                & \textbf{0.9930} & 0.0045 & 0.5004 & 0.0090 \\
CNNSpot               & 0.8200 & 0.4786 & 0.6499 & 0.5767 \\
NPR                   & 0.8610 & 0.1147 & 0.4891 & 0.1829 \\
AIDE                  & 0.7495 & 0.5083 & 0.6293 & 0.5775 \\
\bottomrule
\end{tabular}
\vspace{-8pt}
\end{table}

\begin{table}[t]
  \caption{Accuracy on other OoD test sets.}
  \label{tab:ood_results}
  \centering
  \scriptsize
  \setlength{\tabcolsep}{3pt}
  \begin{tabular}{@{}lccccccc@{}}
    \toprule
    Test Set & \textbf{Defake-o3} & FakeVLM & LEGION & FakeShield & CNNSpot & NPR & AIDE \\
    \midrule
    AIGI-Now          & \textbf{0.9180} & 0.8596 & 0.4997 & 0.8307 & 0.7730 & 0.6925 & 0.8473 \\
    EvalGEN           & \textbf{0.9872} & 0.9668 & 0.0907 & 0.9513 & 0.7660 & 0.2430 & 0.7045 \\
    MNW               & \textbf{0.8871} & 0.8417 & 0.0265 & 0.5072 & 0.6435 & 0.3775 & 0.7712 \\
    \bottomrule
  \end{tabular}
  \vspace{-10pt}
\end{table}

\subsection{Results on GroundFake Test Set}
Table~\ref{tab:groundfake_results} reports classification (Acc/F1), evidence text overlap (BLEU-1/2 and ROUGE-L), and evidence region overlap (IoU) of state-of-the-art AIGI detectors and Defake-o3 on the GroundFake test set. RL consistently improves Acc/F1 and IoU for all three variants. Enabled by interactive visual search for verifiable evidence, Defake-o3 achieves the strongest overall explainable performance, matching the best Acc/F1 while attaining the highest BLEU-1, BLEU-2, ROUGE-L, and IoU.

\subsection{FakeFrontier Benchmark}
Beyond the in-domain evaluation on GroundFake, we further assess out-of-distribution generalization on recent generators using \textbf{FakeFrontier}. FakeFrontier contains 4,000 images, evenly split between real and fake. The real subset is drawn from Open Images V7~\cite{openimagev7} and Chameleon~\cite{aide}, while the fake subset is collected from outputs of 10 recent image generators, including Nano Banana~\cite{nanobanana}, GPT-Image-1.5~\cite{gptimagebyopenai}, and Seedream 4.5~\cite{seedream45}. The Open Images V7 images used in FakeFrontier do not overlap with those used in GroundFake. Unlike GroundFake, FakeFrontier does not provide human evidence annotations and is therefore used to evaluate both classification on the latest generators and explanation quality without relying on reference boxes or texts.
To evaluate explanations under this annotation-free setting, we adopt an MLLM-based protocol on 200 fake images randomly sampled from FakeFrontier. We instantiate the judge with three open-source MLLMs, Qwen3-VL-235B-A22B-Thinking~\cite{qwen3vl}, Kimi K2.5~\cite{kimi}, and GLM-4.6V~\cite{glm}, to reduce evaluator-specific bias. Each evidence item is assessed independently rather than as part of a bundled explanation. We consider two complementary aspects: \textit{evidence quality}, which measures whether an evidence item is visually grounded and specific to AI generation, and \textit{evidence persuasiveness}, which measures whether it can convince the judge to predict ``fake''. Table~\ref{tab:fakefrontier_mllm} reports the MLLM-based evaluation, and Table~\ref{tab:fakefrontier_cls} reports classification results on the full FakeFrontier benchmark.

\noindent\textbf{Quality Evaluation.} 
We first evaluate each evidence item independently with an MLLM judge. The judge assigns a score from 1 to 10, where 1 denotes evidence that is irrelevant, contradictory, or not specific to AI generation, and 10 denotes evidence that is visually grounded and highly specific to AI generation. We normalize the score to $[0,1]$ and report it as the \textbf{Quality Score (QS)}. For each image, we average the normalized scores over its evidence items, and then average the resulting image scores over the evaluation set.

\noindent\textbf{Persuasion Evaluation.} 
We next evaluate whether a single evidence item is strong enough to support a fake verdict. For each fake image, the judge first predicts from the image alone, which reveals its prior tendency and is reported as \textbf{Prior-Acc} in Table~\ref{tab:fakefrontier_mllm}. We then provide one evidence item from a detector and ask the judge to assess the same image again. Each evidence item is tested independently. Because this decision can be sensitive to prompting, we use three system prompts and average the results. We report two metrics: (1) \textbf{Hit@Img}, the proportion of fake images for which at least one evidence item leads the judge to output ``fake''; and (2) \textbf{Hit@Evi}, the proportion of evidence items that lead the judge to output ``fake''. For detectors that output plain text rather than structured JSON, we first use Qwen3-VL-8B-Instruct to split their outputs into independent evidence items.
These metrics are complementary: QS measures the visual grounding and artifact specificity of the evidence, Hit@Img measures whether a detector can surface at least one decisive flaw for a fake image, and Hit@Evi measures how often its individual evidence items are actually useful rather than noisy or hallucinated. As shown in Tables~\ref{tab:fakefrontier_mllm} and~\ref{tab:fakefrontier_cls}, Defake-o3 achieves the highest Acc/F1 on FakeFrontier and also ranks first in QS, Hit@Img, and Hit@Evi under all three judges. Although baseline methods produce more evidence items on average, their much lower Hit@Evi suggests that many of those claims are weak or unpersuasive.

\subsection{Results on External OoD Benchmarks}


Beyond FakeFrontier, we further evaluate classification generalization on three external OoD benchmarks: AIGI-Now~\cite{aiginow}, EvalGEN~\cite{dda}, and MNW~\cite{mnw}.
For these external benchmarks, we report classification accuracy only. To maintain a strict OoD setting, we remove samples from generators that overlap with the GroundFake training set.
As shown in Table~\ref{tab:ood_results}, Defake-o3 achieves the best accuracy on all three datasets, reaching 0.9180 on AIGI-Now, 0.9872 on EvalGEN, and 0.8871 on MNW. It consistently outperforms both explainable MLLM baselines and black-box detectors, showing that the advantage of Defake-o3 is not limited to GroundFake or FakeFrontier. These results suggest that interactive visual search and verifier-guided evidence alignment improve robustness under broader distribution shifts, without sacrificing classification strength.

\subsection{Ablation Studies}
We conduct ablation studies to assess the effectiveness of interactive visual search and the role of the Evidence Verifier.



\noindent\textbf{Effectiveness of Interactive Visual Search:} 
Based on the results in Tables~\ref{tab:groundfake_results}--\ref{tab:fakefrontier_cls}, Defake-o3 exhibits clear advantages in both classification performance and explanation quality over the Defake-CoT model (which lacks the zoom-in mechanism) and the Defake-Direct model (which lacks any reasoning process).

\begin{figure}[t]
  \centering
  \includegraphics[width=0.8\linewidth]{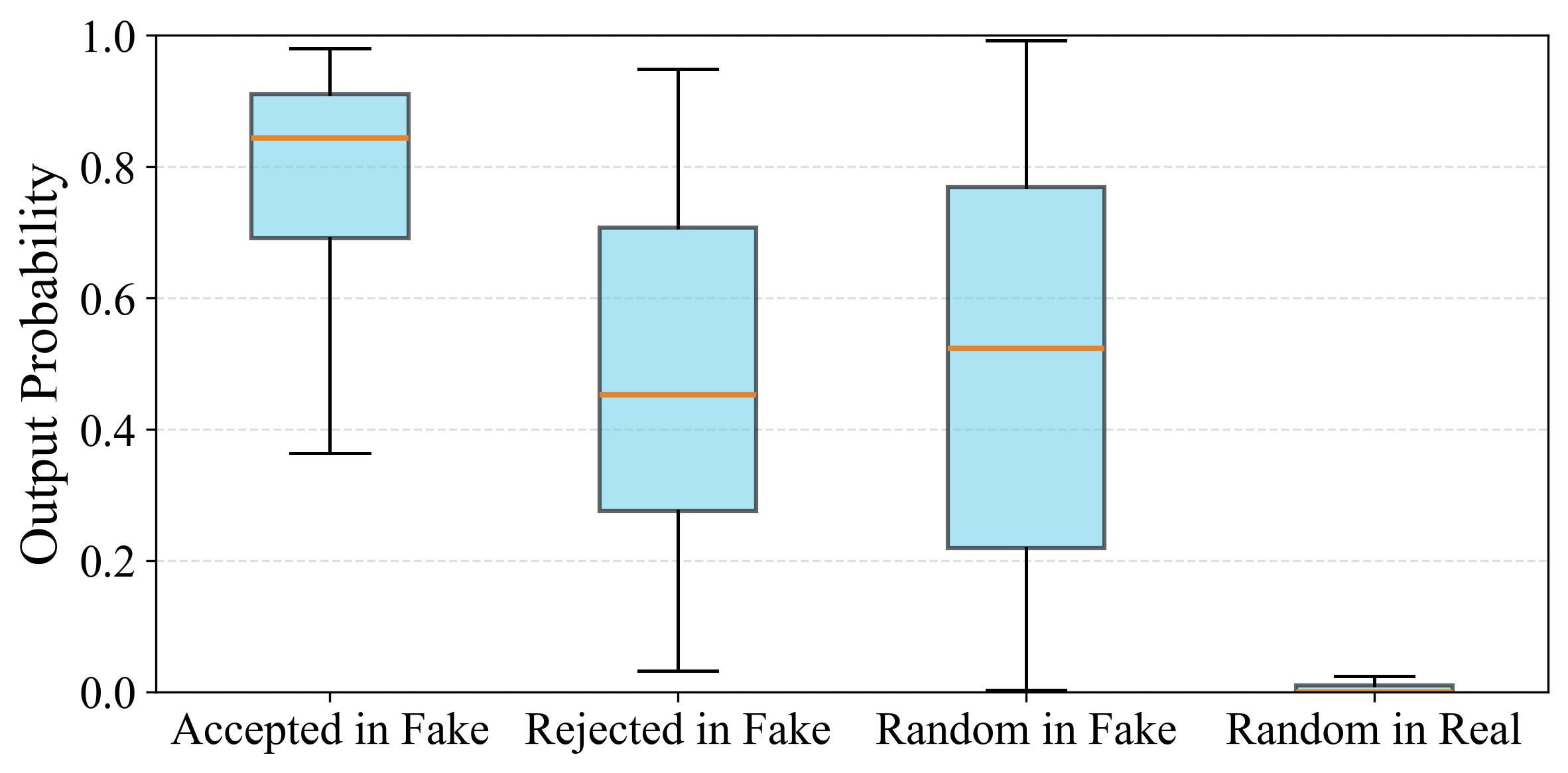}
  \caption{Output value distribution of the Evidence Verifier on generated evidence.}
  \Description{A distribution plot of the Evidence Verifier scores assigned to generated evidence, illustrating the separation between evidence judged invalid and evidence judged valid.}
  \label{fig:verifier_dist}
\end{figure}

\noindent\textbf{The Role of the Evidence Verifier:} 
Figure~\ref{fig:verifier_dist} shows that the Evidence Verifier assigns high scores to human-accepted evidence and much lower scores to human-rejected evidence or forcibly fabricated evidence on random bounding boxes (Random in Fake/Real), indicating that it captures the visual grounding and artifact specificity criteria in our annotation protocol. This learned signal leads to better reward shaping during RL. 
Compared with the rule-only variant ($\alpha_M=0$ in Table~\ref{tab:reward_ablation}), the model trained with the Evidence Verifier improves IoU on GroundFake from 0.252 to 0.311, while causing only minor changes in lexical overlap with the reference annotations.
Table~\ref{tab:reward_ablation} further shows that the best results require a balance between rule-based matching and verifier guidance.
Using only the verifier ($\alpha_M=1$) degrades text-explanation performance.
Although a smaller verifier weight ($\alpha_M=0.25$) yields a slightly higher IoU, it noticeably degrades Acc/F1. We therefore use $\alpha_M=0.5$ as the best overall trade-off. 

\begin{table}[t]
  \caption{Ablation of reward hyperparameters evaluated on the GroundFake test set.}
  \label{tab:reward_ablation}
  \centering
  \scriptsize
  \begin{tabular}{@{}lccccccc@{}}
    \toprule
    $\alpha_M$ & $\lambda_{iou}/\lambda_{bleu}$ & Acc & F1 & BLEU-1 & BLEU-2 & ROUGE-L & IoU \\
    \midrule
    1    & 0 & 0.984 & 0.984 & 0.275 & 0.150 & 0.222 & 0.195 \\
    \midrule
    0.75 & 1  & 0.910 & 0.902 & 0.303 & 0.176 & 0.235 & 0.260 \\
    0.5  & 1  & \textbf{0.992} & \textbf{0.992} & \textbf{0.364} & 0.215 & 0.281 & 0.311 \\
    0.25 & 1  & 0.962 & 0.960 & 0.330 & 0.194 & 0.270 & \textbf{0.318} \\
    0    & 1  & 0.986 & 0.986 & 0.363 & \textbf{0.218} & \textbf{0.291} & 0.252 \\
    \bottomrule
  \end{tabular}

\end{table}


\subsection{Qualitative Results}


Figure~\ref{fig:showcases} illustrates how Defake-o3 turns exploratory inspection into compact, verifiable evidence. It zooms into candidate regions, retains only verifiable artifacts---such as the malformed headset logo, invalid timestamp, and implausible desk object---and discards unsupported suspicions. In contrast, FakeVLM and FakeShield provide coarser evidence that is less directly tied to image-specific artifacts.

\section{Conclusion}
In this paper, we presented Defake-o3, an explainable AIGI detector that moves from speculative rationales to verifiable evidence through interactive visual search and verifier-guided evidence alignment. We further introduced GroundFake and FakeFrontier for grounded training and out-of-distribution evaluation. Experiments show that Defake-o3 achieves the best overall performance among compared methods, with top or tied-for-top classification results and stronger judge-based evidence quality.

\begin{acks}
This work was supported in part by the National Natural Science Foundation of China (Grant Nos. 62302295, 62595733, and 62561160155), the Shanghai Municipal Science and Technology Major Project (Grant No. 2021SHZDZX0102). This work was also supported by Ant Group. In particular, we sincerely thank Zijuan Yu, Yan Hong, and Jun Lan from Ant Group for their valuable help.
\end{acks}

\bibliographystyle{ACM-Reference-Format}
\balance
\bibliography{main}

\clearpage
\appendix
\twocolumn[
\begin{center}
  {\Huge\sffamily\bfseries
    Defake-o3: From Speculative Rationales to Verifiable Evidence for
    Explainable AIGI Detection\par}
  \vspace{0.55em}
  {\huge\sffamily\bfseries Supplementary Material\par}
  \vspace{0.55em}
\end{center}
\vspace{0.7em}
]

\setcounter{figure}{0}
\setcounter{table}{0}
\renewcommand{\thefigure}{S\arabic{figure}}
\renewcommand{\thetable}{S\arabic{table}}

\section{Prompts}
\label{sec:prompts_summary}

In this section, we summarize all the prompts utilized throughout our study. The full content of each prompt is provided at the end of this supplementary material.

\begin{itemize}[leftmargin=*]
    \item \textbf{Prompts for Inference:} The system prompts used during inference by Defake-o3, Defake-CoT, and Defake-Direct are detailed in Figures~\ref{fig:prompt_defake_o3}, \ref{fig:prompt_defake_cot}, and \ref{fig:prompt_defake_direct}, respectively, and the unified user prompt is provided in Figure~\ref{fig:prompt_inference_user}.
    
    \item \textbf{Prompts for GroundFake Dataset Construction:} The system and user prompts used to instruct Gemini 3 Pro Preview to generate evidence candidates are shown in Figures~\ref{fig:prompt_candidate_gen} and \ref{fig:prompt_candidate_user}, and the prompt for trajectory rewriting is detailed in Figure~\ref{fig:prompt_traj_rewrite}.
    
    \item \textbf{Prompts for FakeFrontier Benchmark:} The prompt used for extracting descriptions for image generation is shown in Figure~\ref{fig:prompt_image_captioning}, the prompt for evidence quality evaluation is detailed in Figure~\ref{fig:prompt_quality_eval}, the three system prompts for persuasion evaluation are provided in Figures~\ref{fig:prompt_persuasion_1}, \ref{fig:prompt_persuasion_2}, and \ref{fig:prompt_persuasion_3}, and the prompt for parsing baseline explanations is shown in Figure~\ref{fig:prompt_evidence_parsing}.
\end{itemize}

\section{Robustness Evaluation}

We evaluate the performance of our model on the FakeFrontier benchmark under various image degradations, including JPEG compression (Quality Factor = 70), Gaussian blur ($\sigma = 1$), and resizing ($\times 0.5$). As shown in Table~\ref{tab:robustness}, Defake-o3 exhibits strong robustness against these common visual perturbations.

\section{Discussion on Training Dynamics}

\begin{figure}[h]
  \centering
  \includegraphics[width=0.8\linewidth]{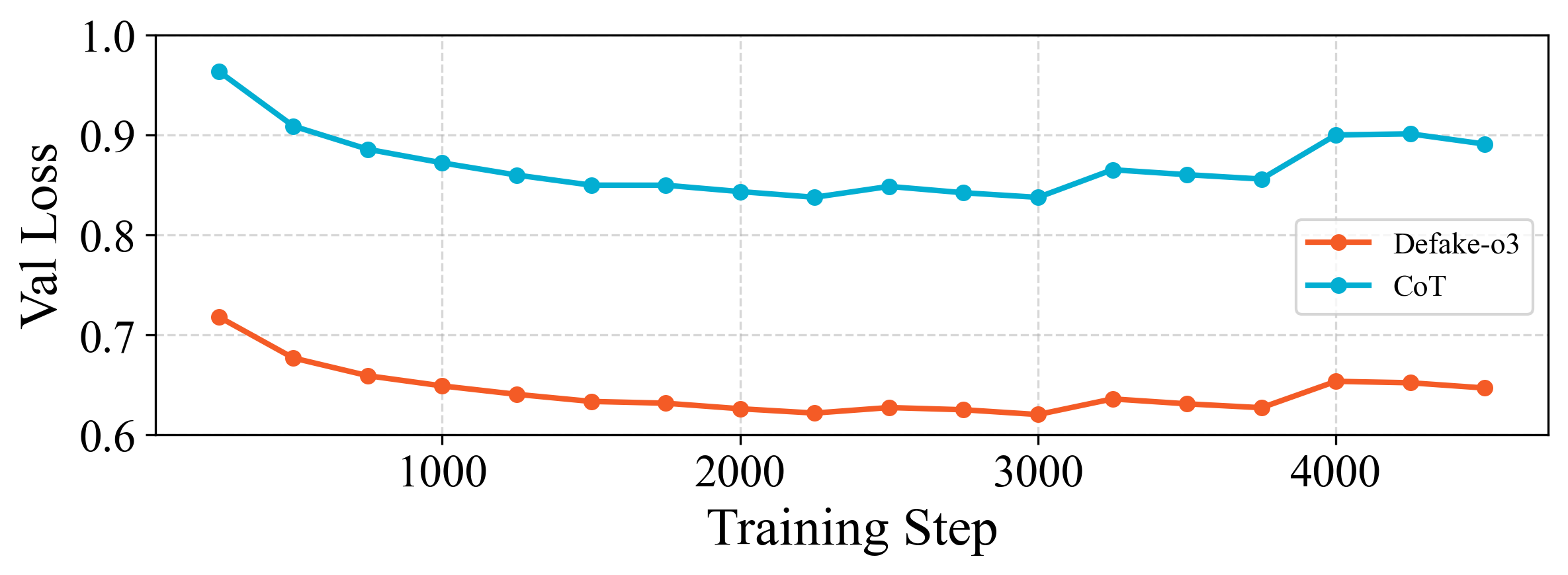}
  \caption{Comparison of validation loss between Defake-o3 and Defake-CoT in the SFT stage.}
  \label{fig:supp_sft}
\end{figure}

During the SFT stage, we consistently observed that the evaluation loss of the Defake-CoT model was significantly higher than that of Defake-o3 (see Figure~\ref{fig:supp_sft}), despite their reasoning text being identical. This indicates that the absence of magnified visual patches makes the reasoning process substantially more difficult.

\begin{figure}[h]
  \centering
  \includegraphics[width=0.8\linewidth]{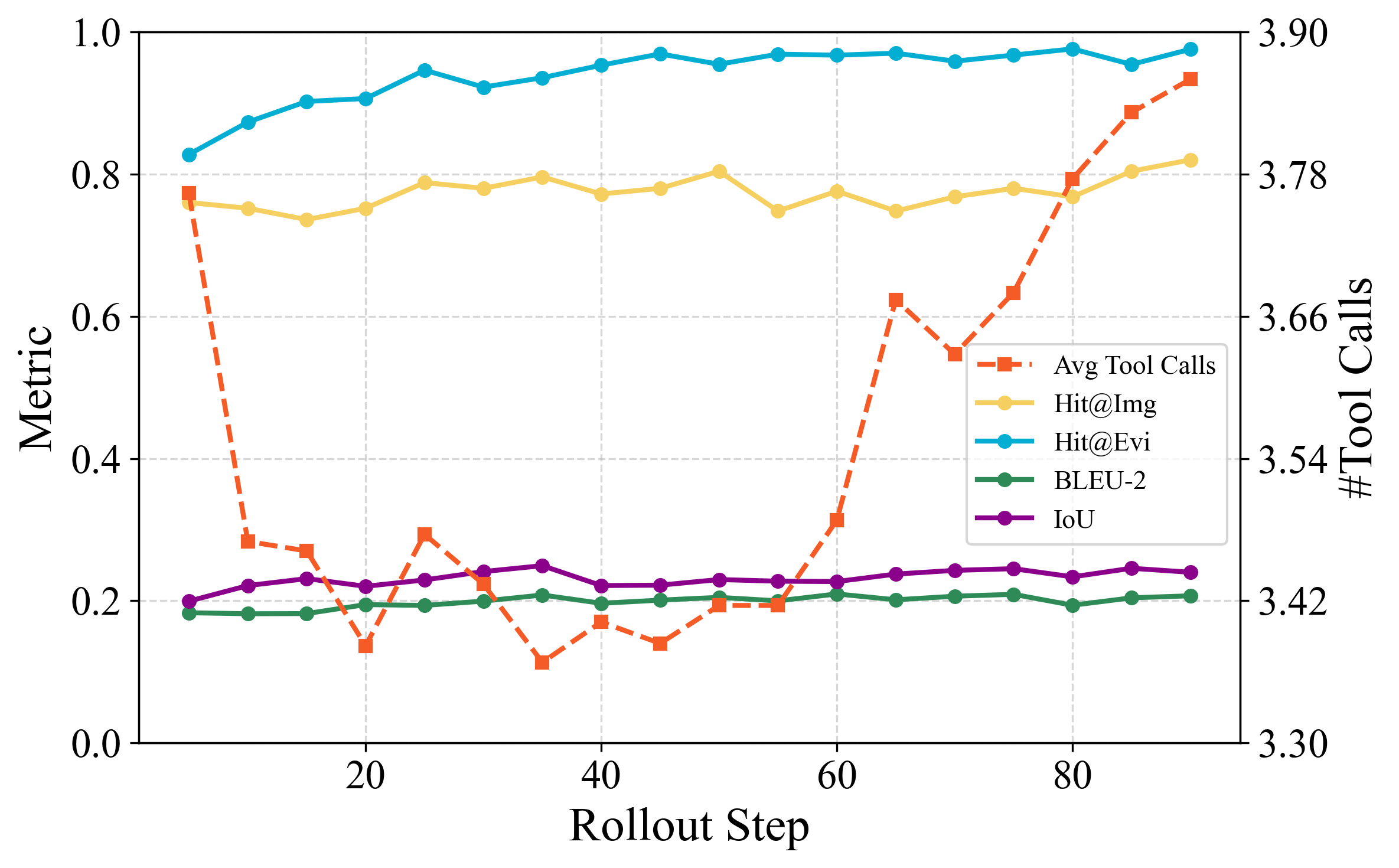}
  \caption{Average number of tool calls and other metrics during the RL stage of Defake-o3.}
  \label{fig:supp_rl}
\end{figure}

Furthermore, during the RL training of Defake-o3, we observed that although we did not introduce any explicit reward for tool invocation, the number of zoom-in tool calls gradually increased after an initial decrease (see Figure~\ref{fig:supp_rl}). This suggests that the model, during the learning process, autonomously overcame the inertia of not invoking tools~\cite{zheng2025deepeyes} and recognized the crucial role of the zoom-in tool in localizing generative flaws.

\section{Ablation on Rule-based Reward}

On the GroundFake test set, incorporating the Evidence Verifier into the reward function significantly improves the Intersection over Union (IoU) compared to the model relying solely on the rule-based reward, albeit with a slight decrease in textual metrics. A natural question arises: if we solely use the rule-based reward but increase the weight of the IoU component, can we achieve the same effect?

As shown in Table~\ref{tab:supp_rule_reward}, increasing the ratio $\lambda_{iou}/\lambda_{bleu}$ under the rule-based reward only setting ($\alpha_M=0$) can improve the IoU to some extent. However, it still falls short of the spatial accuracy achieved when the Evidence Verifier is introduced ($\alpha_M=0.5$). This confirms that the Evidence Verifier provides unique signals beyond what simple bounding box matching can offer.

\section{Base Model Performance}

We report the performance of the pre-trained Qwen3-VL-8B-Instruct base model on the GroundFake dataset without any fine-tuning. During inference, we utilize the same prompt as Defake-Direct. The performances of Defake-Direct and Defake-o3 are also provided in Table~\ref{tab:base_model} for comparison. As can be seen, our training pipeline brings substantial performance improvements over the general-purpose base model.

\section{Inference Speed}

We present a throughput comparison of the Defake-o3, Defake-CoT, and Defake-Direct methods, alongside three explainable baselines (FakeVLM, LEGION, and FakeShield) on an 8$\times$A100 GPU platform. As shown in Table~\ref{tab:inference_speed}, Defake-o3 trades lower inference efficiency for higher detection performance and explanation quality, largely due to the sequential generation and multi-turn interaction nature of its ``thinking-with-images'' paradigm.

\begin{table}[t]
  \caption{Robustness evaluation on FakeFrontier.}
  \label{tab:robustness}
  \centering
  \small
  \begin{tabular}{lcccc}
    \toprule
    Perturbation & Defake-o3 & FakeVLM & LEGION & FakeShield \\
    \midrule
    Original & \textbf{0.9232} & 0.7114 & 0.5004 & 0.6127 \\
    JPEG Compression & \textbf{0.8924} & 0.6891 & 0.5004 & 0.5952 \\
    Gaussian Blur & \textbf{0.8746} & 0.6888 & 0.4991 & 0.6090 \\
    Resize & \textbf{0.9182} & 0.6682 & 0.5004 & 0.5867 \\
    \bottomrule
  \end{tabular}
\end{table}

\begin{table}[t]
  \caption{Ablation of rule-based reward hyperparameters evaluated on the GroundFake test set.}
  \label{tab:supp_rule_reward}
  \centering
  \small
  \begin{tabular}{@{}lccccccc@{}}
    \toprule
    $\alpha_M$ & $\lambda_{iou}/\lambda_{bleu}$ & Acc & F1 & BLEU-1 & BLEU-2 & ROUGE-L & IoU \\
    \midrule
    0.5  & 1  & \textbf{0.992} & \textbf{0.992} & \textbf{0.364} & 0.215 & 0.281 & \textbf{0.311} \\
    \midrule
    0    & 1  & 0.986 & 0.986 & 0.363 & \textbf{0.218} & \textbf{0.291} & 0.252 \\
    0    & 2  & 0.952 & 0.950 & 0.321 & 0.190 & 0.266 & 0.237 \\
    0    & 4  & 0.974 & 0.973 & 0.341 & 0.202 & 0.286 & 0.258 \\
    \bottomrule
  \end{tabular}
\end{table}

\begin{table}[t]
  \caption{Performance of the base model compared to our trained models on GroundFake.}
  \label{tab:base_model}
  \centering
  \scriptsize
  \begin{tabular}{lcccccc}
    \toprule
    Model & Acc & F1 & BLEU-1 & BLEU-2 & ROUGE-L & IoU \\
    \midrule
    Qwen3-VL-8B-Instruct & 0.824 & 0.823 & 0.108 & 0.051 & 0.159 & 0.143 \\
    Defake-Direct & 0.958 & 0.957 & 0.333 & 0.195 & 0.252 & 0.262 \\
    Defake-o3 & \textbf{0.992} & \textbf{0.992} & \textbf{0.364} & \textbf{0.215} & \textbf{0.281} & \textbf{0.311} \\
    \bottomrule
  \end{tabular}
\end{table}

\begin{table}[t]
  \caption{Throughput comparison during inference.}
  \label{tab:inference_speed}
  \centering
  \small
  \begin{tabular}{lc}
    \toprule
    Method & Throughput (samples/second) \\
    \midrule
    Defake-Direct & \textbf{10.668} \\
    Defake-CoT & 8.180 \\
    Defake-o3 & 2.606 \\
    \midrule
    FakeVLM & 3.008 \\
    LEGION & 7.398 \\
    FakeShield & 0.535 \\
    \bottomrule
  \end{tabular}
\end{table}

\section{More Details About GroundFake}

\begin{figure*}[t]
  \centering
  \includegraphics[width=\linewidth]{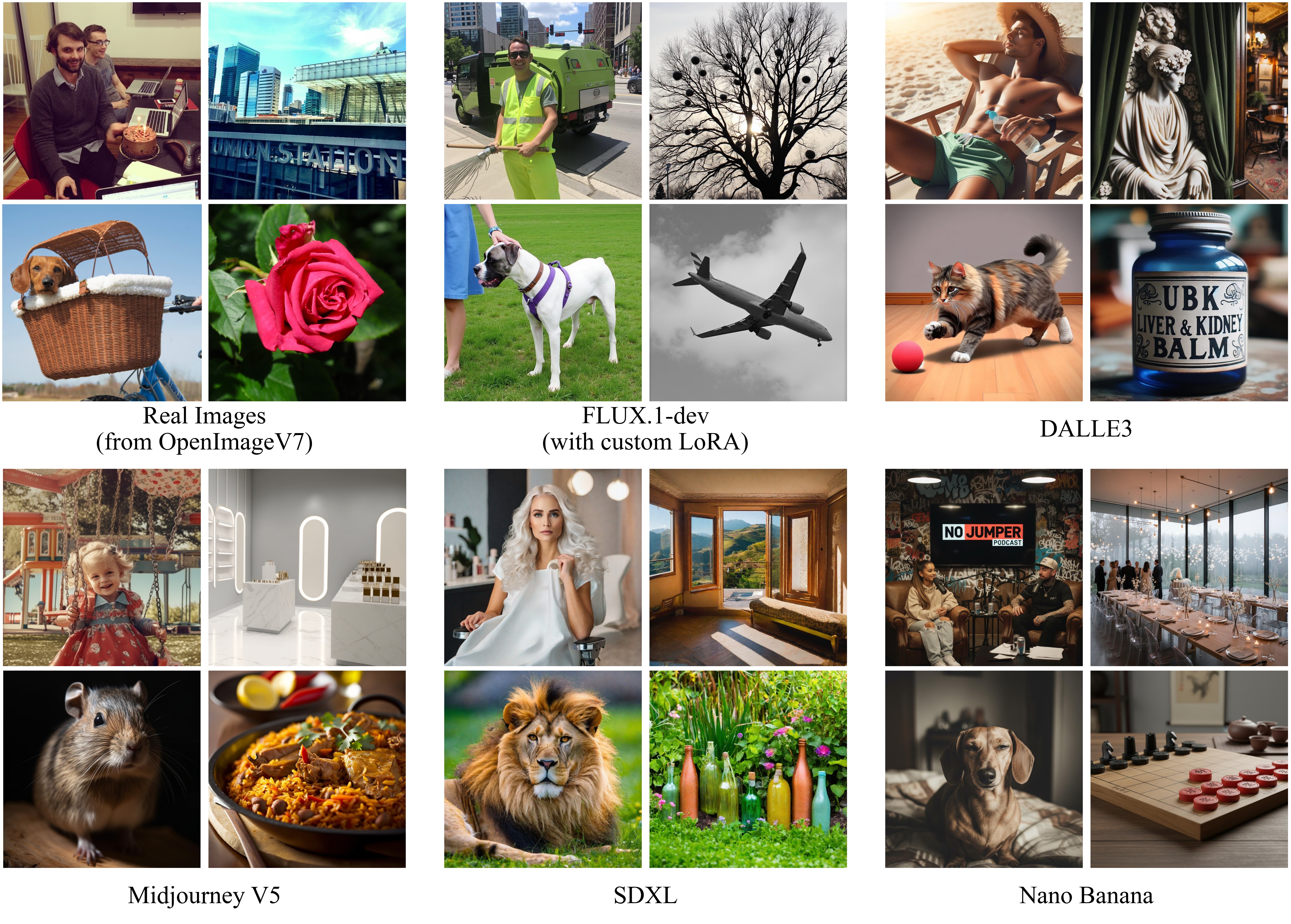}
  \caption{Examples of images from the GroundFake dataset.}
  \label{fig:supp_groundfake}
\end{figure*}

\subsection{Image Collection and Bias Control}

We collected a balanced training set of 16k images (8k real and 8k fake) and applied several bias control procedures to reduce obvious shortcut differences between real and fake images.

\noindent\textbf{Data Sources.} Our training objective is to enable the model to identify subtle generative flaws within AI-generated images. However, if the quality of the generated images in the training set is too low and the errors are overly apparent, the resulting model may perform poorly on high-quality images produced by newer generators. Therefore, we aim to construct a generated image dataset that maintains a high average quality while presenting a gradient of difficulty. To achieve this, we scraped 6.5k generated images from the Internet, covering newer generators including SDXL~\cite{sdxl}, DALLE3~\cite{dalle3}, Midjourney V5~\cite{midjourney}, and Nano Banana~\cite{nanobanana}. For authentic images, we sampled 8k real images from OpenImageV7~\cite{openimagev7}.

Concurrently, we observed that recent generators typically produce images of high aesthetic quality (e.g., perfect composition and lighting). This tendency creates a stylistic discrepancy between the fake image subset and the real image subset, potentially inducing a tendency for the model to overfit to aesthetic features rather than genuine generative artifacts. To counteract this, we generated an additional 1.5k fake images using FLUX.1-dev~\cite{flux1dev} paired with a specialized custom LoRA~\cite{lora}. This LoRA enables the model to simulate the diverse and often imperfect shooting conditions typical of authentic photographs, such as underexposure and motion blur. During this generation process, we utilized the captions of real images from OpenImageV7 as prompts, which effectively produced fake images that closely mimic the diverse styles of real photographs (see Figure~\ref{fig:supp_groundfake}). The final dataset comprises 8k generated images and 8k authentic images. All these images have a longer side of at least 1024 pixels.

\noindent\textbf{Bias Control Procedures.} AIGI detectors can easily rely on dataset-specific shortcuts (non-semantic biases, e.g., aesthetic style) instead of actual generation artifacts. We therefore applied the following procedures:

\begin{itemize}[leftmargin=*]
  \item \textbf{Format Standardization:} All images were converted to high-quality JPG to remove trivial file format cues across different sources.  
  \item \textbf{Aspect Ratio Alignment:} Among the collected fake images, square formats are overrepresented. To reduce the shortcut that square images are fake, we cropped fake images to follow the aspect ratio distribution of OpenImageV7.
  \item \textbf{Category Alignment:} We used Qwen3-VL-8B-Thinking~\cite{qwen3vl} as a coarse classifier to group images into four broad categories: \textit{human}, \textit{animal}, \textit{object}, and \textit{scene}. At the same time, we filtered out clearly non-photorealistic content such as cartoons and digital art. The four categories were kept balanced at a ratio of 1:1:1:1.
  \item \textbf{Aesthetic Quality Alignment:} As mentioned above, recent generators often produce highly polished images, while real photographs span a wider quality range. To reduce the shortcut that very high aesthetic quality implies fake, we matched the aesthetic score distribution of the real subset to that of the fake subset using the LAION-Aesthetic-Predictor~\cite{laionaestheticpredictor}. The additional FLUX.1-dev samples further broaden the fake subset with more diverse photographic conditions, including motion blur, noise, and imperfect lighting.
\end{itemize}

\subsection{Label-Conditioned Candidate Annotation}

To guide the MLLM in generating high-quality explanations, we summarized several common AI generation errors that serve as strong, definitive evidence of synthetic origin. These include: (1) unrecognizable or garbled text; (2) missing, redundant, or distorted object components, including deformed faces or hands; (3) repetitive patterns, such as two identical faces or a single person holding two identical cups simultaneously; (4) anomalous lighting, such as areas appearing unnaturally bright without a clear light source; and (5) other violations of common sense or physical laws, such as objects floating in mid-air. We explicitly incorporated this information into the system prompt to make the model more attentive to these specific issues.

For each image, we employed Gemini 3 Pro Preview~\cite{gemini3pro} to generate reasoning trajectories and evidence candidates. To improve annotation efficiency, we provided the ground-truth label of the image in the prompt. The system prompt used is detailed in Figure~\ref{fig:prompt_candidate_gen} and the user prompt is shown in Figure~\ref{fig:prompt_candidate_user}.

It is worth noting that we simultaneously instructed the model to generate Chinese explanatory text for the evidence to facilitate the annotation process, as all our human annotators are native Chinese speakers.

\subsection{Human Annotation and Verification}

In this stage, the evidence candidates generated for the fake images in the previous step were further inspected and filtered by human annotators. This verification is necessary because the evidence generated by MLLMs frequently exhibits several issues: (1) The descriptive text is entirely irrelevant to the region enclosed by the bounding box; (2) The text contradicts the actual visual content within the bounding box (e.g., text or hands that are rendered correctly are falsely claimed to be deformed); (3) The evidence is insufficient or overly sensitive to be deemed a definitive generative flaw (e.g., claiming ``the character's skin is too smooth,'' which frequently occurs in real images due to post-processing). Meanwhile, the human-annotated data obtained at this stage was also used to train our Evidence Verifier, which learned how to identify such invalid evidence.

During the annotation process, each piece of evidence was marked as either ``Valid'' or ``Invalid''. Annotators were instructed that a piece of evidence should only be marked as ``Valid'' if the text accurately matches the image region within the bounding box AND correctly points out an error that is exclusive to AI-generated images (i.e., not an artifact that could reasonably appear in real photographs due to lighting constraints, simple post-processing, or motion blur).

The annotation of each evidence item was strictly independent. Annotators were provided with the image, the bounding box, and the text in both English and Chinese. A total of 6 annotators participated in this process, with each piece of evidence independently evaluated by 3 of them. To ensure annotation quality, we organized the tasks into batches of approximately 1,000 evidence items. In each batch, we interspersed a certain proportion of evidence from real images, as well as forcibly generated bogus evidence on real images. We mandated that the ``Valid'' rate for these control items must not exceed 5\% within any batch; otherwise, the entire batch was rejected and re-annotated.

Excluding the control items from real images used for quality checks, there was an average of 2.397 pieces of evidence to be annotated per fake image. Under a majority voting rule, 74.99\% of the evidence candidates were ultimately marked as ``Valid''. Furthermore, all 3 annotators reached a unanimous consensus on 89.43\% of the evidence items.

\subsection{Reasoning Trajectory Rewriting for Annotation Consistency}

In this phase, we utilized Gemini 3 Flash Preview~\cite{geminiflash} to rewrite portions of the reasoning trajectories. This step ensures that the thought processes within the trajectories remain logically consistent with the filtered evidence (i.e., after the `Invalid` evidence was removed by human annotators). The system prompt used for this task is provided in Figure~\ref{fig:prompt_traj_rewrite}.

In the user prompt for this task, we provided the original image, the uncorrected output from Gemini 3 Pro Preview, and the human annotation results presented as a boolean list.

\subsection{Impact of Human Verification on Classification Accuracy}

During the human annotation phase, evidence that did not align with human judgment was removed. The primary goal of this procedure was to align the model's explanations with human cognitive standards. Another intriguing question is: does this filtering process also affect the model's overall classification accuracy?

\begin{table}[t]
  \caption{Impact of removing invalid evidence during training.}
  \label{tab:supp_human_verif}
  \centering
  \small
  \resizebox{\linewidth}{!}{
  \begin{tabular}{ccccccc}
  \toprule
  \multirow{2}{*}{Invalid Filtering} 
  & \multirow{2}{*}{RL Training} 
  & \multicolumn{2}{c}{GroundFake} 
  & \multicolumn{3}{c}{FakeFrontier} \\
  \cmidrule(lr){3-4} \cmidrule(lr){5-7}
  & & Acc & IoU & Real Acc & Fake Acc & Acc \\
  \midrule
  \xmark & \xmark & 0.984 & 0.201 & 0.836 & 0.916 & 0.876 \\
  \cmark & \xmark & 0.972 & 0.241 & 0.883 & 0.904 & 0.894 \\
  \cmark & \cmark & \textbf{0.992} & \textbf{0.311} & \textbf{0.902} & \textbf{0.945} & \textbf{0.923} \\
  \bottomrule
\end{tabular}
  }
\end{table}

As shown in Table~\ref{tab:supp_human_verif}, removing human-annotated invalid evidence during the SFT stage understandably results in a higher IoU on the GroundFake test set, because the ground truth references have also been stripped of these invalid elements. Interestingly, this filtering also leads to an increase in overall accuracy on the OoD FakeFrontier benchmark. This improvement stems from a noticeable reduction in the false positive rate on real images. This demonstrates that removing invalid, overly sensitive, or hallucinated evidence during training fundamentally helps reduce the probability of the model misclassifying real images.

\section{More Details About FakeFrontier}

\subsection{Image Sources and Generation Methods}

The 2,000 real images in the FakeFrontier benchmark are evenly sourced from OpenImageV7~\cite{openimagev7} and Chameleon~\cite{aide}. The 2,000 fake images are generated by 10 state-of-the-art generators: GPT-Image-1.5~\cite{gptimagebyopenai}, GPT-Image-1~\cite{gptimage1}, Seedream 4.5~\cite{seedream45}, Seedream 3.0~\cite{seedream30}, Z-Image-Turbo~\cite{zimage}, Qwen-Image~\cite{qwenimage}, Nano Banana~\cite{nanobanana}, FLUX.2-pro~\cite{flux2pro}, HunyuanImage 3.0~\cite{hunyuanimage30}, and Stable Diffusion 3.5 Large~\cite{sd35large}. For each generator, we produced 200 images. Specifically, 100 images were generated using prompts derived from the captions of OpenImagesV7 samples, and the remaining 100 were generated using prompts based on Chameleon captions. We utilized Qwen3-VL-8B-Instruct to generate these captions, instructing the model to preserve the stylistic information (e.g., underexposure, motion blur) of the original real images. This approach ensures that the generated images mimic the diverse aesthetic conditions of real-world photography. The prompt used for generating image captions is detailed in Figure~\ref{fig:prompt_image_captioning}.

During the image generation process, we employed several standard resolutions: $1344\times 768$, $1216\times 832$, $1024\times 1024$, $832\times 1216$, and $768\times 1216$. For each reference real image, its synthetic counterpart was generated using the standard resolution that most closely matched its aspect ratio. All other generation parameters were set to the models' default or officially recommended configurations.

\subsection{MLLM-based Protocol}

We deployed Qwen3-VL-235B-A22B-Thinking, Kimi K2.5, and GLM-4.6V locally to evaluate the explanation quality. During inference, all models used their default generation parameters.

For the Quality Evaluation, we used the system prompt detailed in Figure~\ref{fig:prompt_quality_eval}. For the Persuasion Evaluation, since the susceptibility of Large Language Models to persuasion is highly dependent on the system prompt, we employed three distinct sets of system prompts. The final results presented in the main text are the average scores across these three prompts. Prompt I is detailed in Figure~\ref{fig:prompt_persuasion_1}. The distinction in Prompt II (detailed in Figure~\ref{fig:prompt_persuasion_2}) is the inclusion of common AI generation errors as background knowledge. The distinction in Prompt III (detailed in Figure~\ref{fig:prompt_persuasion_3}) is a warning to the model that the provided evidence might be misleading, forcing it to rely more heavily on its autonomous judgment of the image content. This significantly increases the model's skepticism; consequently, evidence that still manages to persuade the model under these conditions possesses remarkably high quality.

\subsection{Detailed Results of Persuasion Evaluation}

We provide the raw persuasion evaluation results under each of the three individual prompts for reference, as shown in Tables~\ref{tab:persuasion_1}, \ref{tab:persuasion_2}, and \ref{tab:persuasion_3}. It can be observed that the difficulty of persuading the MLLM to deliver a ``fake'' verdict progressively increases from Prompt I to Prompt III. This indicates that evidence capable of persuading the MLLM under the more stringent conditions of Prompt III is significantly more effective and reasonable. Moreover, across all different prompt configurations, Defake-o3 consistently maintains a clear advantage over the other baseline methods.

\begin{table}[t]
  \caption{Persuasion Evaluation Results under Prompt I.}
  \label{tab:persuasion_1}
  \centering
  \small
  \resizebox{\linewidth}{!}{
  \begin{tabular}{l cc cc cc}
    \toprule
    \multirow{2}{*}{Method}
    & \multicolumn{2}{c}{Qwen3-VL-235B}
    & \multicolumn{2}{c}{Kimi K2.5}
    & \multicolumn{2}{c}{GLM-4.6V} \\
    \cmidrule(lr){2-3} \cmidrule(lr){4-5} \cmidrule(lr){6-7}
    & \makebox[1.2cm][c]{Hit@Img} & \makebox[1.2cm][c]{Hit@Evi}
    & \makebox[1.2cm][c]{Hit@Img} & \makebox[1.2cm][c]{Hit@Evi}
    & \makebox[1.2cm][c]{Hit@Img} & \makebox[1.2cm][c]{Hit@Evi} \\
    \midrule
    Defake-o3      & 0.9347 & \textbf{0.8825} & 0.8643 & 0.7292 & 0.9397 & \textbf{0.9527} \\
    Defake-CoT     & \textbf{0.9548} & 0.8684 & \textbf{0.8844} & \textbf{0.7521} & \textbf{0.9548} & 0.8974 \\
    Defake-Direct  & 0.7538 & 0.7165 & 0.6683 & 0.5890 & 0.7688 & 0.6803 \\
    FakeVLM        & 0.5895 & 0.4740 & 0.5474 & 0.4870 & 0.7053 & 0.7428 \\
    FakeShield     & 0.3166 & 0.5420 & 0.3116 & 0.5465 & 0.3266 & 0.5646 \\
    \bottomrule
  \end{tabular}
  }
\end{table}

\begin{table}[t]
  \caption{Persuasion Evaluation Results under Prompt II.}
  \label{tab:persuasion_2}
  \centering
  \small
  \resizebox{\linewidth}{!}{
  \begin{tabular}{l cc cc cc}
    \toprule
    \multirow{2}{*}{Method}
    & \multicolumn{2}{c}{Qwen3-VL-235B}
    & \multicolumn{2}{c}{Kimi K2.5}
    & \multicolumn{2}{c}{GLM-4.6V} \\
    \cmidrule(lr){2-3} \cmidrule(lr){4-5} \cmidrule(lr){6-7}
    & \makebox[1.2cm][c]{Hit@Img} & \makebox[1.2cm][c]{Hit@Evi}
    & \makebox[1.2cm][c]{Hit@Img} & \makebox[1.2cm][c]{Hit@Evi}
    & \makebox[1.2cm][c]{Hit@Img} & \makebox[1.2cm][c]{Hit@Evi} \\
    \midrule
    Defake-o3      & 0.8945 & 0.7455 & 0.8392 & 0.7080 & 0.9497 & \textbf{0.9462} \\
    Defake-CoT     & \textbf{0.9146} & \textbf{0.7607} & \textbf{0.8693} & \textbf{0.7197} & \textbf{0.9598} & 0.8923 \\
    Defake-Direct  & 0.6935 & 0.5543 & 0.6482 & 0.5449 & 0.7688 & 0.6787 \\
    FakeVLM        & 0.4158 & 0.2384 & 0.5316 & 0.4176 & 0.6632 & 0.6561 \\
    FakeShield     & 0.2764 & 0.3741 & 0.3116 & 0.4807 & 0.3266 & 0.5692 \\
    \bottomrule
  \end{tabular}
  }
\end{table}

\begin{table}[t]
  \caption{Persuasion Evaluation Results under Prompt III.}
  \label{tab:persuasion_3}
  \centering
  \small
  \resizebox{\linewidth}{!}{
  \begin{tabular}{l cc cc cc}
    \toprule
    \multirow{2}{*}{Method}
    & \multicolumn{2}{c}{Qwen3-VL-235B}
    & \multicolumn{2}{c}{Kimi K2.5}
    & \multicolumn{2}{c}{GLM-4.6V} \\
    \cmidrule(lr){2-3} \cmidrule(lr){4-5} \cmidrule(lr){6-7}
    & \makebox[1.2cm][c]{Hit@Img} & \makebox[1.2cm][c]{Hit@Evi}
    & \makebox[1.2cm][c]{Hit@Img} & \makebox[1.2cm][c]{Hit@Evi}
    & \makebox[1.2cm][c]{Hit@Img} & \makebox[1.2cm][c]{Hit@Evi} \\
    \midrule
    Defake-o3      & \textbf{0.5226} & \textbf{0.2806} & \textbf{0.5377} & \textbf{0.3312} & \textbf{0.8241} & \textbf{0.5759} \\
    Defake-CoT     & 0.4573 & 0.2427 & 0.3819 & 0.2393 & 0.7739 & 0.4855 \\
    Defake-Direct  & 0.3668 & 0.1874 & 0.3769 & 0.2173 & 0.5879 & 0.3071 \\
    FakeVLM        & 0.1053 & 0.0549 & 0.2789 & 0.1546 & 0.2684 & 0.1286 \\
    FakeShield     & 0.1005 & 0.0726 & 0.2010 & 0.2381 & 0.1709 & 0.1429 \\
    \bottomrule
  \end{tabular}
  }
\end{table}

\subsection{Parsing Explanatory Text}

Our explanation quality evaluation is conducted at the granularity of individual ``evidence'' items. Defake-o3 naturally outputs structured JSON with evidence separated into a list. However, baseline methods often output an unformatted block of text. Therefore, their explanations must be parsed and split before evaluation. We utilized Qwen3-VL-8B-Instruct with the prompt detailed in Figure~\ref{fig:prompt_evidence_parsing} to achieve this.

\section{Details About Baselines}

Here we briefly introduce each baseline method, along with their training details or the source of their weights used in our experiments.

\begin{itemize}[leftmargin=*]
  \item \textbf{CNNSpot}~\cite{cnnspot} is a standard image classifier trained on real and ProGAN-generated images, with enhanced cross-generator generalization via preprocessing, postprocessing, and data augmentation.
  
  We trained it on GroundFake using the Adam optimizer with a learning rate of $1 \times 10^{-4}$ and a batch size of 64 for 15 epochs.  

  \item \textbf{NPR}~\cite{npr} models neighboring pixel relationships induced by generator up-sampling operations and uses these local structural artifacts as a source-invariant representation for generalizable fake image detection.
  
  We trained it on GroundFake using the Adam optimizer with a learning rate of $2 \times 10^{-4}$ and a batch size of 32 for 10 epochs.

  \item \textbf{AIDE}~\cite{aide} combines CLIP-based global semantic embeddings with DCT-guided high/low-frequency patch sampling and SRM-based noise features to detect AI-generated images using hybrid high-level and low-level cues.
  
  We trained it on GroundFake using the AdamW optimizer with a learning rate of $1 \times 10^{-4}$ and a batch size of 32 for 20 epochs.

  \item \textbf{FakeVLM}~\cite{fakeclue} is a specialized multimodal large language model trained on the FakeClue dataset that performs synthetic image classification and generates natural-language artifact explanations from fine-grained visual clues.
  
  We used the pre-trained weights published on their official GitHub repository.

  \item \textbf{LEGION}~\cite{legion} integrates a global image encoder, an MLLM, a grounding image encoder, and a pixel decoder to jointly perform synthetic image detection, artifact localization, and textual explanation.
  
  We reproduced LEGION on their SynthScars dataset using their official training scripts and default parameters.

  \item \textbf{FakeShield}~\cite{fakeshield} combines a domain tag-guided explainable forgery detection module with a multimodal forgery localization module to predict image authenticity, localize manipulated regions, and explain tampering evidence in language.
  
  We used the pre-trained weights published on their official GitHub repository.
\end{itemize}
\section{More Qualitative Results}

We provide additional qualitative results of Defake-o3 on the GroundFake test set, along with corresponding outputs from FakeVLM and FakeShield for a comprehensive comparison of explanation quality.

\begin{figure*}[p]
  \centering
  \vfill
  \includegraphics[width=0.9\linewidth]{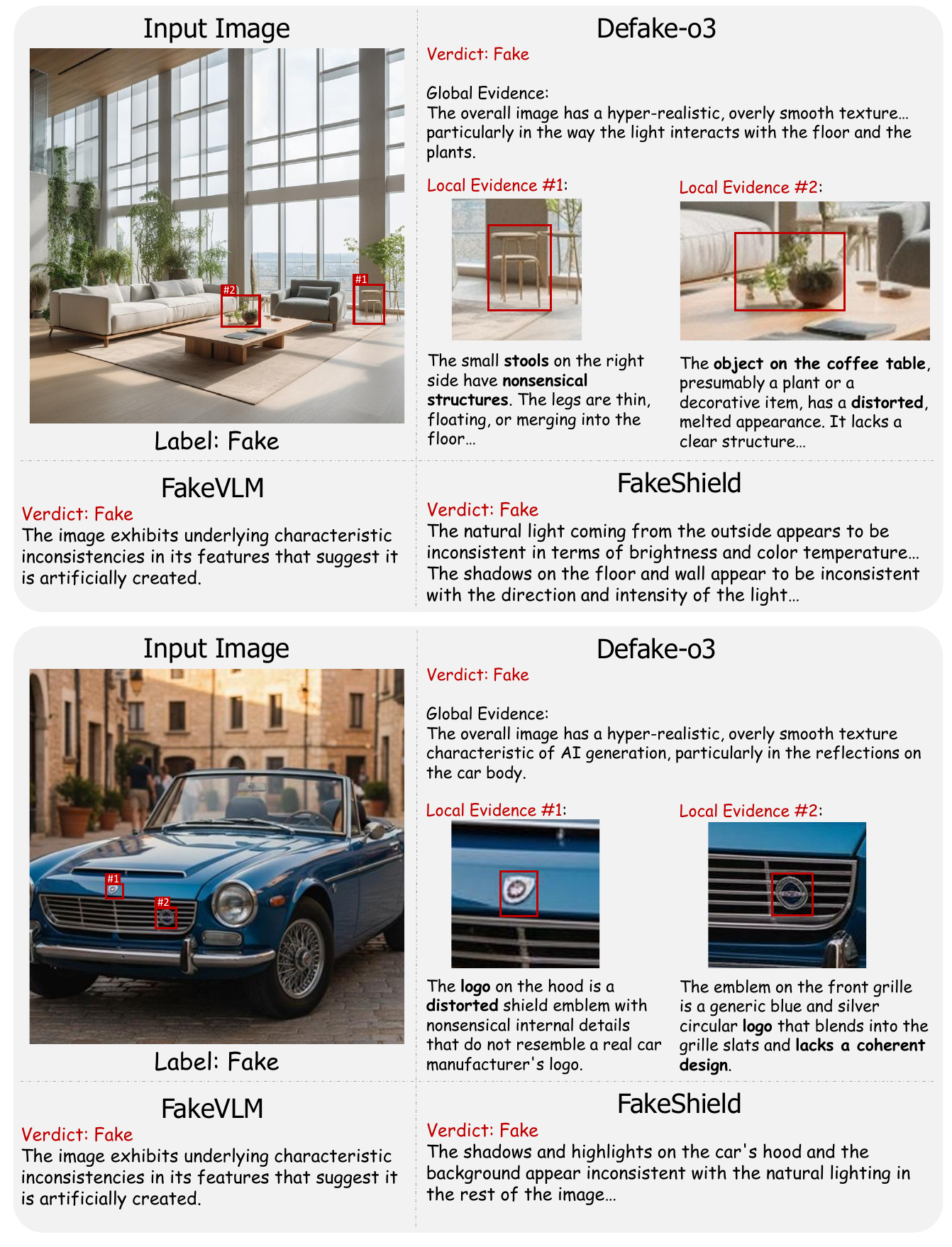}
  \vfill
  \caption{Additional qualitative comparison of Defake-o3 against FakeVLM and FakeShield on the GroundFake test set (Part 1). The bounding boxes are from the outputs of Defake-o3.}
  \label{fig:supp_qualitative_part1}
\end{figure*}

\begin{figure*}[p]
  \centering
  \vfill
  \includegraphics[width=0.9\linewidth]{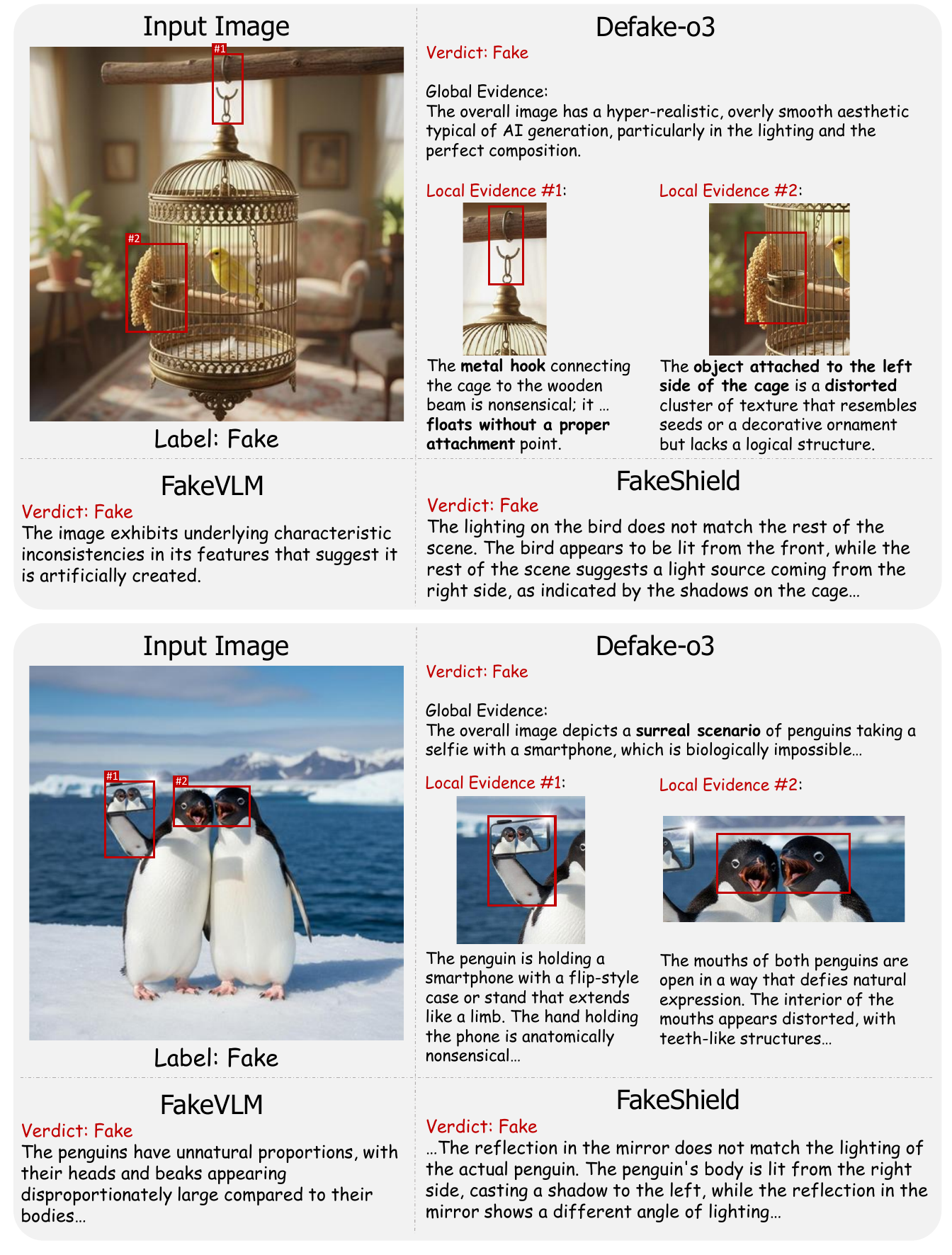}
  \vfill
  \caption{Additional qualitative comparison of Defake-o3 against FakeVLM and FakeShield on the GroundFake test set (Part 2).}
  \label{fig:supp_qualitative_part2}
\end{figure*}

\begin{figure*}[p]
  \centering
  \vfill
  \includegraphics[width=0.9\linewidth]{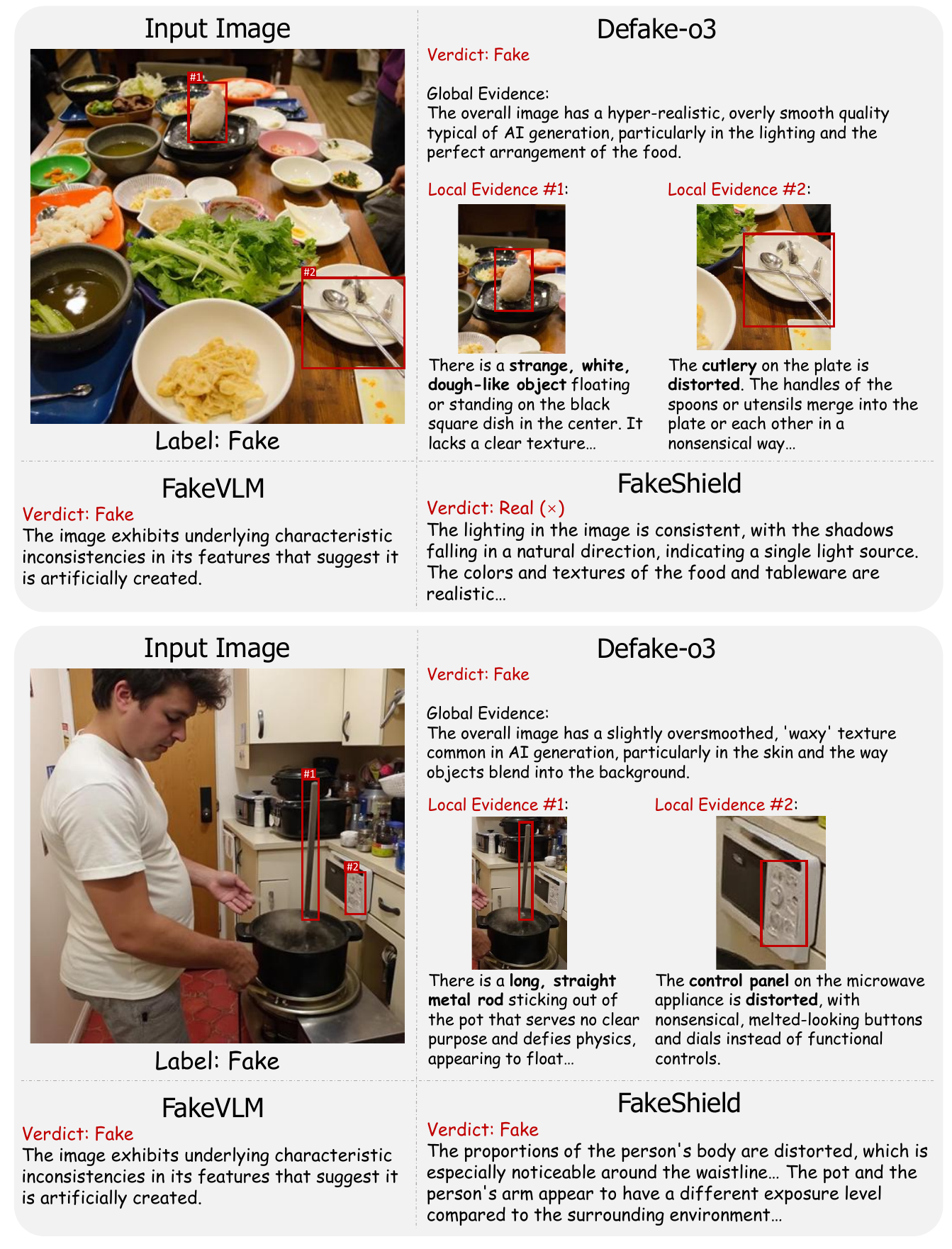}
  \vfill
  \caption{Additional qualitative comparison of Defake-o3 against FakeVLM and FakeShield on the GroundFake test set (Part 3).}
  \label{fig:supp_qualitative_part3}
\end{figure*}

As can be seen from the examples in Figures~\ref{fig:supp_qualitative_part1}, \ref{fig:supp_qualitative_part2}, and \ref{fig:supp_qualitative_part3}, Defake-o3 is able to precisely localize specific generative flaws like unrecognizable objects, and easily identify surreal, obviously AI-generated scenes, such as the two penguins taking a selfie with a smartphone in the bottom example of Figure~\ref{fig:supp_qualitative_part2}. 

In contrast, the explanations provided by the baseline methods are often extremely vague and unreliable. For instance, in 5 out of the 6 examples, FakeVLM outputs the exact same boilerplate explanation: "The image exhibits underlying characteristic inconsistencies in its features that suggest it is artificially created." This indicates that FakeVLM struggles to pinpoint explicit flaws and instead makes unreliable judgments based merely on overall image texture. Meanwhile, FakeShield always attempts to explain its verdicts from the perspective of lighting and shadows, even when the lighting in the generated images does not appear problematic to a human observer. Furthermore, in the bottom example of Figure~\ref{fig:supp_qualitative_part2}, FakeShield mistakenly identifies the smartphone as a "mirror", demonstrating that its fine-tuning process may have compromised the model's pre-existing world knowledge.

\clearpage

\begin{figure*}[!htbp]
\begin{promptbox}
You are an expert in the field of forged image detection.

For reference, here are some common errors found in AI-generated images:
1. Unrecognizable text
2. Missing, redundant, or distorted object components, such as deformed face, deformed limbs, extra or missing limbs, including inconsistency with logos or unique features of well-known brands
3. Repeating patterns, such as two identical faces, one person holding two water cups, or a large number of people with unusually similar appearances
4. Unusual lighting, such as unusually bright areas in the absence of light sources
5. Other obvious anomalies that defy common sense or physics, such as objects floating in mid-air

You will be presented with an image which may be AI generated. Please examine the image carefully, paying particular attention to small details. 
Finally, please provide the following in a JSON file:
1. Your evidence. This should be a list, with each element containing a piece of evidence you found that proves this image is AI-generated or not, consisting of a bounding box (in the format of [x1, y1, x2, y2]) and a piece of text. The bbox can be null, indicating that the evidence is for the entire image. For example, the overall texture of the image, whether it is oversaturated, etc. If the bbox is not null, it indicates that the area is key evidence. Only clear and decisive factors can be considered key evidence. For example, if a small area looks blurry, but in fact the real image may also look blurry in a small area due to image compression, this cannot be considered as critical evidence for AIGC. All your key evidence should be clear and unambiguous. You should always provide a piece of overall evidence with a null bbox, and zero or more pieces of key evidence with bounding boxes. If a piece of evidence requires multiple objects to be compared with each other (for example, two duplicate faces), the bbox should include all of them for direct comparison. If your verdict is "fake", only if you can't find any key evidence should you provide only a piece of general evidence with a null bbox. While your overall evidence can be slightly longer, your key evidence should be concise, just in one or two sentences. If your verdict is "real", just provide an overall evidence with null bbox; otherwise, for "fake" verdict, you can give extra key evidence to support your verdict.
2. Your final verdict, "fake" or "real".

The following shows the json format:
```json
{
	"evidence": [
		{
			"bbox": null or [int, int, int, int], # note: the bbox for the first evidence is always null!
			"text": text
		},
		...
	],
	"verdict": "real" or "fake"
}
```

You should use the image cropping tool to examine each suspicious area of the image one by one. You can only use the tool once per round of conversation.
Think step by step, and put your thoughts within the `<think></think>` tag. Place your tool calls and the JSON text block representing the final result outside the `<think></think>` tag.
In each round of the conversation, you should first think, then use the tool once, or give the final answer. For example, a conversation might look like this:
* <think> The image displays a large stadium, ... I am zooming in on the stadium to check the architectural details. </think> <tool_call>{"name": "crop_image", "arguments": {"x1": 150, "y1": 400, "x2": 780, "y2": 700}}</tool_call>
* tool: here is the cropped image...
* <think> I noticed some text-like features on the right facade of the stadium. I will zoom in to verify if the text is legible or if it consists of pseudo-characters... </think> <tool_call>{"name": "crop_image", "arguments": {"x1": 600, "y1": 500, "x2": 750, "y2": 600}}</tool_call>
* tool: here is the cropped image...
...
* <think> ... I have gathered enough evidence to make my final verdict. </think> ```json ... ```
\end{promptbox}
\caption{System Prompt for Defake-o3.}
\label{fig:prompt_defake_o3}
\end{figure*}

\begin{figure*}[!htbp]
\begin{promptbox}
You are an expert in the field of forged image detection.

For reference, here are some common errors found in AI-generated images:
1. Unrecognizable text
2. Missing, redundant, or distorted object components, such as deformed face, deformed limbs, extra or missing limbs, including inconsistency with logos or unique features of well-known brands
3. Repeating patterns, such as two identical faces, one person holding two water cups, or a large number of people with unusually similar appearances
4. Unusual lighting, such as unusually bright areas in the absence of light sources
5. Other obvious anomalies that defy common sense or physics, such as objects floating in mid-air

You will be presented with an image which may be AI generated. Please examine the image carefully, paying particular attention to small details. 
Finally, please provide the following in a JSON file:
1. Your evidence. This should be a list, with each element containing a piece of evidence you found that proves this image is AI-generated or not, consisting of a bounding box (in the format of [x1, y1, x2, y2]) and a piece of text. The bbox can be null, indicating that the evidence is for the entire image. For example, the overall texture of the image, whether it is oversaturated, etc. If the bbox is not null, it indicates that the area is key evidence. Only clear and decisive factors can be considered key evidence. For example, if a small area looks blurry, but in fact the real image may also look blurry in a small area due to image compression, this cannot be considered as critical evidence for AIGC. All your key evidence should be clear and unambiguous. You should always provide a piece of overall evidence with a null bbox, and zero or more pieces of key evidence with bounding boxes. If a piece of evidence requires multiple objects to be compared with each other (for example, two duplicate faces), the bbox should include all of them for direct comparison. If your verdict is "fake", only if you can't find any key evidence should you provide only a piece of general evidence with a null bbox. While your overall evidence can be slightly longer, your key evidence should be concise, just in one or two sentences. If your verdict is "real", just provide an overall evidence with null bbox; otherwise, for "fake" verdict, give extra key evidence to support your verdict.
2. Your final verdict, "fake" or "real".

The following shows the json format:
```json
{
	"evidence": [
		{
			"bbox": null or [int, int, int, int], # note: the bbox for the first evidence is always null!
			"text": text
		},
		...
	],
	"verdict": "real" or "fake"
}
```

Think step by step, and put your thoughts within the `<think></think>` tag. Place your JSON text block representing the final result outside the `<think></think>` tag.
\end{promptbox}
\caption{System Prompt for Defake-CoT.}
\label{fig:prompt_defake_cot}
\end{figure*}

\begin{figure*}[!htbp]
\begin{promptbox}
You are an expert in the field of forged image detection.

For reference, here are some common errors found in AI-generated images:
1. Unrecognizable text
2. Missing, redundant, or distorted object components, such as deformed face, deformed limbs, extra or missing limbs, including inconsistency with logos or unique features of well-known brands
3. Repeating patterns, such as two identical faces, one person holding two water cups, or a large number of people with unusually similar appearances
4. Unusual lighting, such as unusually bright areas in the absence of light sources
5. Other obvious anomalies that defy common sense or physics, such as objects floating in mid-air

You will be presented with an image which may be AI generated. Please examine the image carefully, paying particular attention to small details. 
Finally, please provide the following in a JSON file:
1. Your evidence. This should be a list, with each element containing a piece of evidence you found that proves this image is AI-generated or not, consisting of a bounding box (in the format of [x1, y1, x2, y2]) and a piece of text. The bbox can be null, indicating that the evidence is for the entire image. For example, the overall texture of the image, whether it is oversaturated, etc. If the bbox is not null, it indicates that the area is key evidence. Only clear and decisive factors can be considered key evidence. For example, if a small area looks blurry, but in fact the real image may also look blurry in a small area due to image compression, this cannot be considered as critical evidence for AIGC. All your key evidence should be clear and unambiguous. You should always provide a piece of overall evidence with a null bbox, and zero or more pieces of key evidence with bounding boxes. If a piece of evidence requires multiple objects to be compared with each other (for example, two duplicate faces), the bbox should include all of them for direct comparison. If your verdict is "fake", only if you can't find any key evidence should you provide only a piece of general evidence with a null bbox. While your overall evidence can be slightly longer, your key evidence should be concise, just in one or two sentences. If your verdict is "real", just provide an overall evidence with null bbox; otherwise, for "fake" verdict, give extra key evidence to support your verdict.
2. Your final verdict, "fake" or "real".

The following shows the json format:
```json
{
	"evidence": [
		{
			"bbox": null or [int, int, int, int], # note: the bbox for the first evidence is always null!
			"text": text
		},
		...
	],
	"verdict": "real" or "fake"
}
```

You should directly give this JSON text block.
\end{promptbox}
\caption{System Prompt for Defake-Direct.}
\label{fig:prompt_defake_direct}
\end{figure*}

\begin{figure*}[!htbp]
\begin{promptbox}
Here is an image which may be AI generated. Please observe carefully and give your verdict and reasoning.
\end{promptbox}
\caption{User Prompt for Inference.}
\label{fig:prompt_inference_user}
\end{figure*}

\begin{figure*}[!htbp]
\begin{promptbox}
You are an expert in the field of forged image detection.

For reference, here are some common errors found in AI-generated images:
1. Unrecognizable text
2. Missing, redundant, or distorted object components, such as deformed face, deformed limbs, extra or missing limbs, including inconsistency with logos or unique features of well-known brands
3. Repeating patterns, such as two identical faces, one person holding two water cups, or a large number of people with unusually similar appearances
4. Unusual lighting, such as unusually bright areas in the absence of light sources
5. Other obvious anomalies that defy common sense or physics, such as objects floating in mid-air

You will be presented with an image which may be AI generated. Please examine the image carefully, paying particular attention to small details. You should zoom in on specific areas for a closer look. Finally, please provide the following in a JSON file:
1. Your exploration trajectory. This should be a list, with each element containing the area you explored and your thought. Each area is given as a bounding box in the format [ymin, xmin, ymax, xmax] normalized to 0-1000. If you explored the entire image, the bbox should be null. At the end of each of your thoughts, you should mention which area you plan to explore next, or state that you are going to present your final results. You should explore the image thoroughly, always starting with the whole image (a null bbox), and then zoom in to up to 5 areas.
2. Your evidence. This should also be a list, with each element containing a piece of evidence you found that proves this image is AI-generated or not, also consisting of a bounding box and a piece of text. The bbox can be null, indicating that the evidence is for the entire image. For example, the overall texture of the image, whether it is oversaturated, etc. If the bbox is not null, it indicates that the area is key evidence. Only clear and decisive factors can be considered key evidence. For example, if a small area looks blurry, but in fact the real image may also look blurry in a small area due to image compression, this cannot be considered as critical evidence for AIGC. All your key evidence should be clear and unambiguous. You should always provide a piece of overall evidence with a null bbox, and zero or more pieces of key evidence with bounding boxes. If a piece of evidence requires multiple objects to be compared with each other (for example, two duplicate faces), the bbox should include all of them for direct comparison. If your verdict is "fake", only if you can't find any key evidence should you provide only a piece of general evidence with a null bbox. While your overall evidence can be slightly longer, your key evidence should be concise, just in one or two sentences. If your verdict is "real", just provide an overall evidence with null bbox; otherwise, for "fake" verdict, give extra key evidence to support your verdict. Additional requirement: Please provide a Chinese translation of the text.
3. Your final verdict, "fake" or "real".

The following shows the json format:
```json
{
	"exploration_trajectory": [
		{
			"bbox": null or [int, int, int, int],   # note: the bbox for the first exploration step is always null!
			"thought": text
		},
		...
	],
	"evidence": [
		{
			"bbox": null or [int, int, int, int], # note: the bbox for the first evidence is always null!
			"text": text,
			"chinese_text": text
		},
		...
	],
	"verdict": "real" or "fake"
}
```
\end{promptbox}
\caption{System Prompt for Evidence Candidate Generation (GroundFake).}
\label{fig:prompt_candidate_gen}
\end{figure*}

\begin{figure*}[!htbp]
\begin{promptbox}
Here is an image which may be AI generated. Please observe carefully and give your verdict and reasoning.       
Hint: The correct verdict should be '<label>'. But please behave as if you don't know this hint before you give your final verdict.
\end{promptbox}
\caption{User Prompt for Evidence Candidate Generation (GroundFake).}
\label{fig:prompt_candidate_user}
\end{figure*}

\begin{figure*}[!htbp]
\begin{promptbox}
Here is a prompt for evaluating AI-generated images with a large language model:

---
You are an expert in the field of forged image detection.

For reference, here are some common errors found in AI-generated images:
1. Unrecognizable text
2. Missing, redundant, or distorted object components, such as deformed face, deformed limbs, extra or missing limbs, including inconsistency with logos or unique features of well-known brands
3. Repeating patterns, such as two identical faces, one person holding two water cups, or a large number of people with unusually similar appearances
4. Unusual lighting, such as unusually bright areas in the absence of light sources
5. Other obvious anomalies that defy common sense or physics, such as objects floating in mid-air

You will be presented with an image which may be AI generated. Please examine the image carefully, paying particular attention to small details. You should zoom in on specific areas for a closer look. Finally, please provide the following in a JSON file:
1. Your exploration trajectory. This should be a list, with each element containing the area you explored and your thought. Each area is given as a bounding box in the format [ymin, xmin, ymax, xmax] normalized to 0-1000. If you explored the entire image, the bbox should be null. At the end of each of your thoughts, you should mention which area you plan to explore next, or state that you are going to present your final results. You should explore the image thoroughly, always starting with the whole image (a null bbox), and then zoom in to up to 5 areas.
2. Your evidence. This should also be a list, with each element containing a piece of evidence you found that proves this image is AI-generated or not, also consisting of a bounding box and a piece of text. The bbox can be null, indicating that the evidence is for the entire image. For example, the overall texture of the image, whether it is oversaturated, etc. If the bbox is not null, it indicates that the area is key evidence. Only clear and decisive factors can be considered key evidence. For example, if a small area looks blurry, but in fact the real image may also look blurry in a small area due to image compression, this cannot be considered as critical evidence for AIGC. All your key evidence should be clear and unambiguous. You should always provide a piece of overall evidence with a null bbox, and zero or more pieces of key evidence with bounding boxes. If a piece of evidence requires multiple objects to be compared with each other (for example, two duplicate faces), the bbox should include all of them for direct comparison. If your verdict is "fake", only if you can't find any key evidence should you provide only a piece of general evidence with a null bbox. While your overall evidence can be slightly longer, your key evidence should be concise, just in one or two sentences. If your verdict is "real", just provide an overall evidence with null bbox; otherwise, for "fake" verdict, give extra key evidence to support your verdict. Additional requirement: Please provide a Chinese translation of the text.
3. Your final verdict, "fake" or "real".

The following shows the json format:
```json
{
        "exploration_trajectory": [
                {
                        "bbox": null or [int, int, int, int],   # note: the bbox for the first exploration step is always null!
                        "thought": text
                },
                ...
        ],
        "evidence": [
                {
                        "bbox": null or [int, int, int, int], # note: the bbox for the first evidence is always null!
                        "text": text,
                        "chinese_text": text
                },
                ...
        ],
        "verdict": "real" or "fake"
}
```
---

Note that this prompt is not for you. Instead, here is already the output from using this prompt with an image.  You will see the image, the analysis result and a bool list. The analysis result was obtained using the above prompt. And the bool list is of the same length of the evidence list, indicating whether each item of the evidence is correct. Your goal is to correct the analysis result based on the bool list.

For each item of the bool list, if it is true or null, you should leave the corresponding item of the evidence as it is.
If it is false, it means that this is not key evidence of an AI-generated image according to expert evaluation. 
And you should remove this item from the evidence list. At the same time, you should update the thought in the exploration trajectory related to this evidence item, no longer considering it as decisive evidence. Also, you may or may not need to update the first evidence item (with null bbox) because this is the overall evidence for the entire image.

Here are some examples:
* Evidence: "The text on the red box is unrecognizable"; Marked as: False; Reason: The text is not clearly visible at the current resolution, regardless of whether it is a real image or an AI-generated image.
* Evidence: "The person's hands are deformed"; Marked as: False; Reason: The person's hands look normal from the current angle.

By the way, you should keep "verdict" unchanged. You should give the corrected analysis result in the same json format as above.
\end{promptbox}
\caption{System Prompt for Trajectory Rewriting (GroundFake).}
\label{fig:prompt_traj_rewrite}
\end{figure*}

\begin{figure*}[!htbp]
\begin{promptbox}
You are an expert photo captioner. Describe the photo in natural language using at most two sentences. Include: 
1. A concise description of the image's style or shooting conditions in one sentence. For example, "Professional photography", "Photograph with natural lighting and high depth of field", "Amateur photo from the 2000s, in dim lighting, taken with a flash", "Amateur photo, even lighting, casual composition", "Amateur photo, with slight motion blur, handheld camera".
2. The main subjects, their actions, the setting.
Now give your caption directly.
\end{promptbox}
\caption{Prompt for Image Captioning (FakeFrontier).}
\label{fig:prompt_image_captioning}
\end{figure*}

\begin{figure*}[!htbp]
\begin{promptbox}
You will see an image which may be AI generated or not, along with a piece of evidence that attempts to prove that the image was generated by AI.
Your task is not to judge the authenticity of the image, but rather to evaluate the quality of the provided evidence. You should determine whether the evidence is sufficient and convincing enough to support the claim that the image is AI-generated.
You should rate the quality of the evidence on a scale of 1 (worst) to 10 (best)
where 1 means the evidence is weak, irrelevant, or does not match the image at all;
and 10 means the evidence is clear, specific, and directly supports the claim that the image is AI-generated.   
You should explain why you give the rating, and you should give your final rating in a pair of tags like this:  

<rating>1~10</rating>
\end{promptbox}
\caption{System Prompt for Quality Evaluation (FakeFrontier).}
\label{fig:prompt_quality_eval}
\end{figure*}

\begin{figure*}[!htbp]
\begin{promptbox}
You will see an image which may be AI generated or not, along with a piece of evidence that attempts to prove that the image was generated by AI.
Your task is to determine whether the image was generated by AI, referring to this evidence.
Provide your verdict and reasons; your final verdict should be given in a **JSON text block** like this:        
```json
{
    "verdict": "real" or "fake"
}
```
\end{promptbox}
\caption{Persuasion Evaluation Prompt I (FakeFrontier).}
\label{fig:prompt_persuasion_1}
\end{figure*}

\begin{figure*}[!htbp]
\begin{promptbox}
You are an expert in the field of forged image detection.

For reference, here are some common errors found in AI-generated images:
1. Unrecognizable text
2. Missing, redundant, or distorted object components, such as deformed face, deformed limbs, extra or missing limbs, including inconsistency with logos or unique features of well-known brands
3. Repeating patterns, such as two identical faces, one person holding two water cups, or a large number of people with unusually similar appearances
4. Unusual lighting, such as unusually bright areas in the absence of light sources
5. Other obvious anomalies that defy common sense or physics, such as objects floating in mid-air

You will see an image which may be AI generated or not, along with a piece of evidence that attempts to prove that the image was generated by AI.
Your task is to determine whether the image was generated by AI, referring to this evidence.
Provide your verdict and reasons; your final verdict should be given in a **JSON text block** like this:        
```json
{
    "verdict": "real" or "fake"
}
```
\end{promptbox}
\caption{Persuasion Evaluation Prompt II (FakeFrontier).}
\label{fig:prompt_persuasion_2}
\end{figure*}

\begin{figure*}[!htbp]
\begin{promptbox}
You are an expert in the field of forged image detection.

For reference, here are some common errors found in AI-generated images:
1. Unrecognizable text
2. Missing, redundant, or distorted object components, such as deformed face, deformed limbs, extra or missing limbs, including inconsistency with logos or unique features of well-known brands
3. Repeating patterns, such as two identical faces, one person holding two water cups, or a large number of people with unusually similar appearances
4. Unusual lighting, such as unusually bright areas in the absence of light sources
5. Other obvious anomalies that defy common sense or physics, such as objects floating in mid-air

You will see an image which may be AI generated or not, along with a piece of evidence that attempts to prove that the image was generated by AI.
Your task is to determine whether the image was generated by AI, referring to this evidence.
Note that the evidence may be misleading, and you should make a verdict based on the actual content of the image.
Provide your verdict and reasons; your final verdict should be given in a **JSON text block** like this:        
```json
{
    "verdict": "real" or "fake"
}
```
\end{promptbox}
\caption{Persuasion Evaluation Prompt III (FakeFrontier).}
\label{fig:prompt_persuasion_3}
\end{figure*}

\begin{figure*}[!htbp]
\begin{promptbox}
You are an assistant that converts AIGC-detection model output text into JSON.
Return ONLY a JSON object with this schema:
{
  "evidence": [
    {"text": "reason point 1"},
    {"text": "reason point 2"}
  ]
}
Rules:
- Split the reasons into concise key points.
- Keep each point factual and directly based on the input text.
- Do not include bbox or any fields other than evidence[].text.
- Use English.
\end{promptbox}
\caption{Prompt for Evidence Parsing.}
\label{fig:prompt_evidence_parsing}
\end{figure*}

\FloatBarrier

\end{document}